\pdfoutput=1

\documentclass[11pt]{article}

\usepackage[final]{acl}

\usepackage{times}
\usepackage{latexsym}

\usepackage[T1]{fontenc}

\usepackage[utf8]{inputenc}

\usepackage{microtype}

\usepackage{inconsolata}

\usepackage{graphicx}

\usepackage{float}
\usepackage{tabularx} 
\usepackage{booktabs} 
\usepackage{comment} 
\usepackage{wrapfig} 
\usepackage{calligra}

\usepackage{aurical}

\newcommand*{\escape}[1]{\texttt{\textbackslash#1}}
\newcommand*{\escapeI}[1]{\texttt{\expandafter\string\csname #1\endcsname}}

\newcommand\blfootnote[1]{%
  \begingroup
  \renewcommand\thefootnote{}\footnote{#1}%
  \addtocounter{footnote}{-1}%
  \endgroup
}

\makeatletter
\def\thickhline{%
  \noalign{\ifnum0=`}\fi\hrule \@height \thickarrayrulewidth \futurelet
   \reserved@a\@xthickhline}
\def\@xthickhline{\ifx\reserved@a\thickhline
               \vskip\doublerulesep
               \vskip-\thickarrayrulewidth
             \fi
      \ifnum0=`{\fi}}
\makeatother

\makeatletter
\newcommand\footnoteref[1]{\protected@xdef\@thefnmark{\ref{#1}}\@footnotemark}
\makeatother

\newlength{\thickarrayrulewidth}
\title{Is Child-Directed Speech Effective Training Data for Language Models?}

\author{Steven Y. Feng, Noah D. Goodman, Michael C. Frank \\ Stanford University \\ \texttt{\{syfeng,ngoodman,mcfrank\}@stanford.edu}}

\begin{document}

\maketitle

\begin{abstract}
While high-performing language models are typically trained on hundreds of billions of words, human children become fluent language users with a much smaller amount of data. What are the features of the data they receive, and how do these features support language modeling objectives? To investigate this question, we train GPT-2 and RoBERTa models on 29M words of English child-directed speech and a new matched, synthetic dataset (TinyDialogues), comparing to OpenSubtitles, Wikipedia, and a heterogeneous blend of datasets from the BabyLM challenge. We evaluate the syntactic and semantic knowledge of these models using developmentally-inspired evaluations. Through pretraining experiments, we test whether the global developmental ordering or the local discourse ordering of children's training data supports high performance relative to other datasets. The local properties of the data affect model results, but surprisingly, global properties do not. Further, child language input is not uniquely valuable for training language models. These findings support the hypothesis that, rather than proceeding from better data, the child's learning algorithm is substantially more data-efficient than current language modeling techniques. 
\end{abstract}

\section{Introduction}
\label{sec:intro}

\blfootnote{Code \& data: \href{https://github.com/styfeng/TinyDialogues}{https://github.com/styfeng/TinyDialogues}}

Transformer-based language models (LM) show very strong performance on a wide variety of downstream tasks, but typically only after pretraining on hundreds of billions to trillions of words \citep{brown2020language}. In contrast, human learners use language fluently after far less training data -- in the 10s to 100s of millions of words. This ``data gap'' \citep{frank2023bridging} of several orders of magnitude poses a substantial challenge for machine learning. 

Is the source of human children's efficient learning a function of their data or their learning algorithms? While children receive rich multi-modal input from their exploration of the world, here we focus on their language input, which has been a major focus of study in developmental psychology \citep{macwhinney2014childes}. One hypothesis is that the language data that children receive is a uniquely rich learning signal -- conversational interaction with their caregivers -- that is curricularized optimally to support learning \citep{eaves2016infant, you2021child, newport1990maturational}. Indeed, interventions to increase the quality of caregiver language do produce improvements in children's language learning \citep{ferjan2020parent}, and interventions to simplify model training data also result in stronger performance \citep{muckatira2024emergent, eldan2023tinystories}. 

Language model pretraining experiments provide a targeted method for investigating dataset quality \citep{kallini2024mission}: we can manipulate the training data available to models to create ``controlled rearing'' experiments. We take advantage of this method to investigate the properties of child-directed speech for learning the syntactic and semantic structure of language. We use GPT-2 and RoBERTa as our simulated learners, and pretrain on natural and synthetic child-language data. For each of these, we conduct two experiments. First, we investigate whether the natural curricularization of children's input -- from simpler utterances to more complex conversations -- affects language model learning. Second, we test whether the local discourse coherence structure of dialogue results in better learning. Finally, we compare to learning on other data sources: OpenSubtitles (more general conversation data), Wikipedia, and a more heterogeneous blend of data from various sources.

We find that the curricularization of child language does not provide a uniquely valuable signal for language models, supporting the hypothesis that other aspects of children's learning (not simply the data) -- perhaps interactions with their training data -- are responsible for their efficiency relative to language models. On the other hand, 
the source, composition, and local properties of the training data have measurable effects on model performance.



\section{Related Work}
\label{sec:related_work}

The efficiency of children's learning has been an important focal point for recent NLP efforts \citep{huebner2021babyberta, zhang2020you}. Last year's BabyLM challenge held the training data for models constant, while encouraging entrants to investigate alternative learning architectures \citep{warstadt2023findings}. Smaller models of this type must be evaluated using more appropriate targeted benchmarks, including evaluations of semantic \citep{zhuang2023visual} and grammatical abilities \citep{huebner2021babyberta, warstadt-etal-2020-blimp-benchmark}. These evaluations have even been used to benchmark performance based on data from a single child \citep{qin2024systematic}. There have also been multimodal investigations of children's learning, particularly in the form of developmental egocentric video data, e.g. \citet{sullivan2021,long2024babyviewdatasethighresolutionegocentric}.

The method of ``controlled rearing'' (manipulating data while holding the model constant) for language models \citep{frank2023openly} has a long history in cognitive science, e.g. \citet{christiansen1999toward}, but has recently become prominent for testing learnability claims \citep{warstadt2024artificial, kallini2024mission, misra2024language}. Often, models trained on naturally-occurring corpora are contrasted with counterfactual corpora constructed via targeted experimental manipulations -- for example, shuffling sentence ordering \citep{kallini2024mission} or removing particular constructions \cite{misra2024language}.

Curricularization of training data is widely investigated in machine learning \citep{bengio2009curriculum}, with the guiding idea being that an appropriate ordering of training examples can lead to a smoother path to the desired objective. Children's development is argued to create a curriculum to facilitate their learning \citep{smith2018developing, cusack2024helpless}, and \textit{starting small} is hypothesized to be efficient for language learning \citep{NEWPORT1988147,ELMAN199371}. In one study from the visual domain, \citet{sheybani2024curriculum} trained self-supervised models on data from infants and found that a developmental ordering leads to stronger eventual performance compared with a reversed ordering. Our study tests this hypothesis in the language domain.

Language curricularization has been investigated as part of the BabyLM challenge \cite{warstadt2023findings}. Our goal is not to assess curriculum learning more generally but to measure the extent to which the specific developmental curriculum that children are exposed to is helpful. To do this, we focus on the developmental data available to children, i.e. child-directed speech. On the other hand, BabyLM contributors relied on the use of proxies rather than age-related information, including ranking sentences by surprisal \citep{chobey-etal-2023-training,hong-etal-2023-surprisal} and lexical frequency \citep{borazjanizadeh-2023-optimizing,martinez-etal-2023-climb}. 


\section{Methods}
\label{sec:methodology}

\subsection{Datasets}
\label{sec:dataset}

\paragraph{CHILDES} The Child Language Data Exchange System (CHILDES) is a repository of human-transcribed corpora of children and caregivers' talk \citep{macwhinney2014childes}, with children ranging from birth to age 13. We take the English subset, which consists of approximately 29M total words (including speaker labels and other metadata) across $\approx$11k conversations. 
CHILDES is heavily skewed towards younger ages; $\approx$90\% of the data is for children ages 2-5 (see Figure \ref{fig:CHILDES_words_skew_graph} in Appendix \ref{appendix:data_statistics}).

\paragraph{TinyDialogues} Inspired by TinyStories \cite{eldan2023tinystories}, we collect a synthetic dataset consisting of approximately 29M words called TinyDialogues (TD). Using GPT-4, we prompted the generation of realistic conversations involving children of ages 2, 5, 10, and 15 years as the central participant, along with a list of other potential participants (e.g. mom, teacher, babysitter). To diversify, we seeded each conversation based on a list of words known by children at the relevant age and varied the conversation type and length (see Appendix \ref{appendix:TD_collection}).

\paragraph{BabyLM} We further compare to the dataset distributed by the BabyLM challenge \citep{warstadt2023findings}, a 100M word dataset that is a mixture of several sources including transcribed speech, child-directed speech (e.g. CHILDES), children's storybooks, and Wikipedia. It is designed to approximate the language data that a 10-year-old child could receive. We sub-sampled $\approx$29M words from BabyLM to match the size of our other data.

\paragraph{Wikipedia} We further compare to Wikipedia data, which is a comprehensive, crowd-sourced online encyclopedia, containing formal and expository text on diverse topics. 
We take a mixture of the Wikipedia and Simple Wikipedia subsets of the BabyLM data, where the latter is a simplified version with shorter sentences and simpler vocabulary. We sub-sampled $\approx$29M total words.

\paragraph{OpenSubtitles} We also compare to OpenSubtitles, which contains more general conversation data in the form of movie and TV subtitles. We sub-sampled $\approx$29M total words.

\paragraph{Preprocessing} Training data for CHILDES and TD was set up so that each line corresponded to a single conversation. Training data for BabyLM, Wikipedia, and OpenSubtitles was set up using the pre-existing format in the BabyLM challenge, 
which mainly consisted of examples spanning across multiple lines. Each dataset was then split into 85/15 train/val splits, of approximately 24.5M training words and 4.5M validation words. 

We include the child's speech in our training data. This is consistent with previous work, e.g. \citet{huebner2021babyberta} in BabyLM. Our goal is to assess the properties of child-directed speech as training input. Such speech in real households contains both the input to the child and the child's responses. Removing the child's speech would create incoherent training data that lacked context. Another approach, as in \citet{huebner2021babyberta}, would be to remove speaker labels entirely. However, the child's utterances come from a different distribution, and this decreases language modeling performance relative to including speaker labels -- see Table \ref{tab:eval_results_speaker-labels} in Appendix \ref{appendix:speaker_labels}. Hence, we choose to include the child’s utterances with speaker labels. 
We append a speaker label for each evaluation example, as we found this more effective (see Appendix \ref{appendix:zorro}).



\subsection{Evaluation}

\paragraph{Zorro} \citep{huebner2021babyberta} is designed for child-directed language and aims to quantify the syntactic and grammatical knowledge of language models. It does so by assessing their capability to distinguish between minimal pairs of sentences that exhibit various grammatical contrasts. We report final averages (of accuracy, higher is better) across individual Zorro tasks in Section \ref{sec:results_and_analysis}.

\paragraph{Word Similarity} To assess the semantic knowledge of our models, we employ a word similarity (WS) metric \cite{zhuang2023visual}, which measures the ability of models to capture semantic similarities between pairs of words. 
We extract word embedding representations from hidden layers of each model, compute pairwise cosine similarities between these embeddings, and report Spearman correlations between human and model similarity judgments (higher is better). The best layer of each model is chosen. We average results across several word similarity benchmarks including RG-65 \cite{rg-65}, WordSim-353 \cite{wordsim-353}, SimLex-999 \cite{hill-etal-2015-simlex}, SimVerb-3500 \cite{gerz-etal-2016-simverb}, and MEN (MTest-3000) \cite{bruni-etal-2012-distributional}.

\subsection{Experiments}

\paragraph{Global Ordering} To test whether the natural ordering of speech to children presents an effective curriculum for model learning, we ordered our CHILDES and TD training examples in three ways: 1) age order (from younger to older), 2) reverse order (from older to younger), and 3) random order (equivalent to randomly shuffling the training data). CHILDES includes fine-grained age information of the target (main) child involved in each conversation, down to fractions of months (essentially days), and we ordered conversations based on this information. TD was ordered based on the conversation seed ages of 2, 5, 10, and 15 years old. For the random order experiments, we randomly shuffled the conversations and kept this shuffled order for all experiments for consistency purposes.

\paragraph{Local Ordering} To investigate the effects of local ordering on learning, we ordered utterances within each CHILDES and TD conversation in two ways: 1) normal (original) order, 2) random order. The latter breaks the local discourse coherence. 

\subsection{Model Training}
\label{sec:model_training}
We use the autoregressive LM GPT-2 \citep{radford2019language} with 124M parameters (small version), following prior ``controlled rearing'' work \citep{kallini2024mission, misra2024language, qin2024systematic}. 
We also experiment using RoBERTa \citep{liu2019robertarobustlyoptimizedbert} with 125M parameters (base version), a masked language model (MLM) pretrained by predicting what should be the $<mask>$ tokens given past and future context. For both models, we trained a separate tokenizer on each of our datasets, and pretrained GPT-2 and RoBERTa from scratch using a learning rate (LR) of $1e-04$ and $5e-05$, respectively, linear LR scheduler with no warmup, varying batch sizes (4 to 64) per GPU, up to three training seeds (42, 0, 123), and Adam optimizer with $\beta=(0.9,0.999)$ and $\epsilon=1e-08$.


During training, GPT-2 processes data in 1024-token chunks, while RoBERTa uses 512-token chunks. For TD, each conversation is instead treated as a single example padded or truncated to 512 tokens. Most TD conversations fit within this limit, making this effective. In contrast, CHILDES conversations are longer, and truncating them would result in heavy data loss. BabyLM, OpenSubtitles, and Wikipedia examples span multiple lines without clear end-of-example markers.

For our global ordering experiments, we split each dataset into $b$ approximately equal sections (buckets), and trained on each repeatedly ($n$ times) before moving to the next bucket. 
This technique was intended as a compromise between standard techniques for model training -- which require iterated training on a dataset -- and human learning -- which operates via a single pass through ordered training data. 
For TD, we used the data corresponding to the four seed ages as the four \textit{buckets}. For CHILDES, we experimented with different numbers of buckets ($b$) and settled on $b=5$ for most experiments. To compare to BabyLM (which cannot be \textit{bucketed}), we also trained GPT-2 and RoBERTa using the standard iterative training approach on each dataset for 20 and 50 epochs, respectively, selecting the epoch that performed best on the respective validation split (lowest val loss).

\section{Results and Analysis}
\label{sec:results_and_analysis}

\begin{table}[t]
\centering
\scalebox{0.92}{
\begin{tabular}{lcc}
\toprule
\textbf{Model} & \textbf{Zorro} & \textbf{WS}\\
\midrule
CHILDES                   & 78.29\% $\pm$ 0.51\% & 0.24 $\pm$ 0.01\\
TD                      & 78.48\% $\pm$ 0.82\% & 0.42 $\pm$ 0.01\\
Wikipedia              & 78.16\% $\pm$ 0.61\% & 0.32 $\pm$ 0.02\\
OpenSubtitles          & 81.02\% $\pm$ 1.03\% & 0.38 $\pm$ 0.00\\
BabyLM                & 82.90\% $\pm$ 1.01\% & 0.42 $\pm$ 0.01\\
\bottomrule
\end{tabular}
}
\caption{Evaluation results (average and standard deviation across three seeds) of our GPT-2 models across datasets, using standard iterative training for 20 epochs.}
\label{tab:eval_results_datasets}
\end{table}

\begin{table}[t]
\centering
\scalebox{0.92}{
\begin{tabular}{lcc}
\toprule
\textbf{Model} & \textbf{Zorro} & \textbf{WS}\\
\midrule
CHILDES                   & 58.37\% $\pm$ 0.96\% & 0.14 $\pm$ 0.01\\
TD                      & 78.52\% $\pm$ 3.10\% & 0.27 $\pm$ 0.04\\
Wikipedia              & 60.84\% $\pm$ 2.06\% & 0.34 $\pm$ 0.02\\
OpenSubtitles          & 62.57\% $\pm$ 0.62\% & 0.20 $\pm$ 0.03\\
BabyLM                & 59.43\% $\pm$ 4.86\% & 0.30 $\pm$ 0.03\\
\bottomrule
\end{tabular}
}
\caption{Evaluation results (avg. and std. across two seeds) of our RoBERTa models across datasets, using standard iterative training for 50 epochs.}
\label{tab:eval_results_datasets_roberta}
\end{table}

\begin{table}[t]
\centering
\scalebox{0.82}{
\begin{tabular}{lccc}
\toprule
\textbf{Dataset} & \textbf{Order} & \textbf{Zorro} & \textbf{WS}\\
\midrule
CHILDES                 & Age &  75.62\% $\pm$ 1.16\% & 0.20 $\pm$ 0.01\\
CHILDES                 & Reverse &  77.63\% $\pm$ 1.29\% & 0.20 $\pm$ 0.01\\
CHILDES                 & Random & 76.87\% $\pm$ 1.12\% & 0.19 $\pm$ 0.01\\
\midrule
TD                 & Age & 78.16\% $\pm$ 0.11\% & 0.32 $\pm$ 0.01\\
TD                 & Reverse & 77.71\% $\pm$ 0.21\% & 0.32 $\pm$ 0.01\\
TD                 & Random & 79.53\% $\pm$ 2.09\% & 0.34 $\pm$ 0.01\\
\bottomrule
\end{tabular}
}
\caption{Evaluation results (avg. and std. across three seeds) of our GPT-2 models, comparing global ordering methods using the repeated buckets training approach, broken down by dataset. For CHILDES, we use $b=5,n=10$, and for TD, we use $n=10$.
}
\label{tab:eval_results_global_order}
\end{table}

\begin{table}[t]
\centering
\scalebox{0.82}{
\begin{tabular}{lccc}
\toprule
\textbf{Dataset} & \textbf{Order} & \textbf{Zorro} & \textbf{WS}\\
\midrule
CHILDES                 & Normal & 78.29\% $\pm$ 0.51\% & 0.24 $\pm$ 0.01\\
CHILDES                 & Random & 77.34\% $\pm$ 1.02\% & 0.19 $\pm$ 0.01\\
\midrule
TD                 & Normal & 78.48\% $\pm$ 0.82\% & 0.42 $\pm$ 0.01\\
TD                 & Random & 78.38\% $\pm$ 0.79\% & 0.42 $\pm$ 0.00\\
\bottomrule
\end{tabular}
}
\caption{Evaluation results (avg. and std. across three seeds) of our GPT-2 models, comparing local ordering methods, broken down by dataset. We use standard iterative training for 20 epochs.}
\label{tab:eval_results_local_order}
\end{table}

\begin{table}[t]
\centering
\scalebox{0.82}{
\begin{tabular}{lccc}
\toprule
\textbf{Dataset} & \textbf{Order} & \textbf{Zorro} & \textbf{WS}\\
\midrule
CHILDES                 & Normal & 58.37\% $\pm$ 0.96\% & 0.14 $\pm$ 0.01\\
CHILDES                 & Random & 55.96\% $\pm$ 0.13\% & 0.04 $\pm$ 0.01\\
\midrule
TD                 & Normal & 78.52\% $\pm$ 3.10\% & 0.27 $\pm$ 0.04\\
TD                 & Random & 79.30\% $\pm$ 3.35\% & 0.24 $\pm$ 0.04\\
\bottomrule
\end{tabular}
}
\caption{Evaluation results (avg. and std. across two seeds) of our RoBERTa models, comparing local ordering methods, broken down by dataset. We use standard iterative training for 50 epochs.}
\label{tab:eval_results_local_order_roberta}
\end{table}

Major results of our experiments can be found in Tables \ref{tab:eval_results_datasets} to \ref{tab:eval_results_local_order_roberta}. Statistical significances can be found in Tables \ref{tab:eval_results_datasets_$p$-values} to \ref{tab:eval_results_local_order_roberta_$p$-values} in Appendix \ref{appendix:stat_sig}. More detailed results of the curricularization experiments can be found in Appendix \ref{appendix:results}.

As seen in Table \ref{tab:eval_results_datasets}, GPT-2 trained on BabyLM outperforms all other datasets on Zorro (syntax) and WS (semantics). OpenSubtitles also surpasses CHILDES and Wikipedia. TD performs best on WS, tying with BabyLM. This suggests that a diverse mixture of data sources, or more varied conversational data, may be more effective for training smaller autoregressive models on limited data. Additionally, synthetic conversation data appears more effective than natural data for training such models at a smaller scale.

From Table \ref{tab:eval_results_datasets_roberta}, RoBERTa shows different patterns, with TD outperforming other datasets on Zorro, OpenSubtitles ranking second (but much lower on WS), and Wikipedia excelling in WS. Our synthetic conversation data (TD) is effective for MLM learning of syntax and grammar, but less so for semantics. TD and OpenSubtitles' focus on dialogue dynamics may favor syntactic learning but struggle with nuanced semantics, where Wikipedia's diverse, factual content excels, particularly for MLM-based learning of semantics. Overall, conversational data seems essential for better grammar and syntax learning across architectures.


CHILDES continues to perform the worst on both Zorro and WS, and synthetic conversation data proves more effective than natural data for small-scale LM training. CHILDES is heavily skewed towards younger ages (see Figure \ref{fig:CHILDES_words_skew_graph} in Appendix \ref{appendix:data_statistics}), whereas TD is more uniform across ages with more sophisticated conversations intended to simulate speech to older children. As such, it contains a higher fraction of more grammatical utterances. While collecting TD, we ensured that it was diverse in conversation type, participants, and content, likely resulting in a more comprehensive coverage of the distribution of potential conversations. This may lead to more effective learning of syntax and semantics, and similar logic likely applies to OpenSubtitles. Further, high-quality synthetic data -- in contrast to naturalistic data, which contains disfluencies and occasional garbled tokens due to transcription issues -- may simply be better suited for training LMs, especially when data is limited.

While Wikipedia is complex and diverse, it may lack conversational elements crucial for small-scale grammar learning, such as back-and-forth interaction and pragmatic cues found in dialogue. Its expository style also limits exposure to informal speech patterns, which could be important for improving syntactic understanding at smaller scales. 

As seen in Table \ref{tab:eval_results_global_order}, global ordering has a negligible effect on GPT-2 performance, with Zorro and WS results remaining relatively stable across different orderings. This is surprising, as curriculum learning -- starting with simpler utterances and conversations and progressing to more complex ones -- might be expected to enhance model learning, similar to humans, 
Aligning with this, while local training behavior (e.g. loss per epoch) varied with ordering, the high-level behavior of the validation loss remained relatively stable (see Appendix \ref{appendix:convergence_behavior}). This suggests that language models, particularly with limited data, may not benefit from curricularization as much as humans.

We omit RoBERTa global ordering experiment results here as our repeated buckets training approach did not work well; the models seemed unable to converge properly. Their behavior was close to random, barely achieving above chance on Zorro and WS, and we do not interpret them here. Results can be found in Table \ref{tab:eval_results_global_order_roberta} in Appendix \ref{appendix:results}.

From Tables \ref{tab:eval_results_local_order} and \ref{tab:eval_results_local_order_roberta}, we see that local ordering affects model performance. Disrupting discourse coherence negatively affects Zorro and WS for CHILDES, despite Zorro focusing on single-sentence evaluations. The effect, especially on WS, is more pronounced for CHILDES than TD, likely due to CHILDES' shorter average utterances ($\approx$ 4 words vs. 13). Hence, reordering CHILDES utterances likely has a greater effect on the model's ability to learn semantics across a larger set of short utterances. Surprisingly, random utterance order has little to no effect on TD performance, suggesting that TD, and possibly synthetic data, may be more robust to local coherence disruptions.

\section{Conclusion \& Future Work}
\label{sec:conclusion_future_work}


Why do children need much less data than language models to achieve fluency? In experiments with GPT-2 and RoBERTa on CHILDES, OpenSubtitles, Wikipedia, BabyLM, and our synthetic TinyDialogues dataset, we found that synthetic child-directed data outperformed natural child-directed data. In general, more diverse datasets (e.g. general conversation data or a mixture of different data sources) may result in better learning than homogeneous child-directed data. Interestingly, global developmental curricularization had little impact, whereas local discourse coherence mattered, especially for natural child-directed conversation data. In sum, it seems that the curricularization of child language does not provide a uniquely valuable signal for language models. However, the source, composition, and local properties of the training data affect model learning. We hope that future work builds on our work here to expand upon the available evaluation benchmarks and data mixtures for comparison between models and children.

\section*{Limitations}

Some limitations of our work include our current suite of evaluation benchmarks and models. We can expand our benchmarks to include more theory of mind and developmental psychology-inspired benchmarks, and ones for longer coherency evaluation. We can investigate ways to improve curriculum learning with RoBERTa, including alternatives or modifications of the repeated buckets training approach. We can also experiment with larger language models such as LLama-3. Further, we limited our investigations to conversation data, Wikipedia, and the BabyLM mixture. We could explore more types and sources of data, and different varieties and proportions of data mixtures. Additionally, the CHILDES dataset is heavily skewed towards younger ages. To the best of our knowledge, a more balanced and uniform dataset of high-quality textual transcriptions of child-directed conversations is not currently available, but we could consider collecting one in the future. However, this may be less of an issue as Zorro (and many other developmental benchmarks) mainly look at phenomena that are acquired at quite an early age. Overall, these are directions to potentially improve and expand upon our work in the future. We feel that, despite these potential limitations, our current work is an insightful and focused contribution.

\section*{Ethical Considerations}

The majority of our datasets and evaluation benchmarks are already existing, publicly available datasets and benchmarks, intended for public use.

We collected TinyDialogues using GPT-4, following all intended use purposes and OpenAI's policies. Further, the dataset is entirely synthetic, and does not include personal or private information. As a safe and controlled language model, there is an incredibly low risk of offensive content, especially as it involves conversations with younger children. We also manually examined a large subset of the data and ensured there were no ethical issues. This includes profanities, racism, bias, offensive words, and other malicious language.

We acknowledge the potential weaknesses of our trained models, which are small in scale and limited in performance. We will never use or encourage their use for real-world purposes. Our initial experiments are conducted purely for investigation purposes to test our hypotheses. We feel that our work is an important contribution to the ML, NLP, cognitive science, and psychology communities, and we encourage researchers to expand upon it.

Our models, TinyDialogue dataset, and accompanying publication are intended only for research purposes and to assess the effectiveness of child-directed speech for training language models. We do not foresee any explicit way that malicious actors could specifically misuse our trained models or models that could be trained on our dataset.


\section*{Acknowledgments}
This work is funded by a gift from Amazon, a Microsoft Accelerating Foundation Models Research (AFMR) grant, and the NSERC Postgraduate Scholarships – Doctoral (PGS D) program. We are also grateful for additional compute support from MultiOn AI. We would like to thank several folks for their useful insights and feedback including members of the Language and Cognition Lab at Stanford; Alvin Tan, Anjie Cao, and Bobby Sparks, among others. We appreciate insights and support from Kanishk Gandhi, Uri Hasson, Chengxu Zhuang, Yang Liu, Devamanyu Hazarika, and Mahdi Namazifar. Lastly, we thank our ACL Rolling Review (ARR) reviewers and meta-reviewer for their helpful comments and feedback.


\bibliography{acl_latex}

\vspace{5mm}
\appendix
\section{TinyDialogues: Dataset Collection Details \& Examples}
\label{appendix:TD_collection}

Here we discuss some further dataset collection details for TinyDialogues (TD), with examples of TD conversations in Table \ref{tab:TD_examples}.

The specific GPT-4 model we use for collecting our entire dataset is \textit{gpt-4-1106-preview}, which is GPT-4 Turbo with training data up to Apr 2023. 
To increase the diversity of the generated conversations, when prompting GPT-4, we also specify the particular type of conversation (Table \ref{tab:TD_conversation_types}), the approximate length or number of turns (5 or 10),\footnote{GPT-4 had a tendency to generate longer conversations, around 10 and 20 turns instead, respectively.} other potential participants in the conversation (Table \ref{tab:TD_participants}), and certain words (one noun, one verb, and one adjective) sampled from Wordbank CDI \cite{frank2021variability} (ages 2 \& 5) and AoA \cite{Kuperman2012AgeofacquisitionRF} (ages 10 \& 15), cut off by the seeded age, that must be included in the conversation for content diversity. The list of potential participants and content words varied by age, e.g. a 15-year-old teenager would likely not talk regularly with a babysitter. We also collect some additional metadata: a list and description of all participants in the conversation, and a brief description of the context/setting. We only use the dialogue portions for our experiments. The GPT-4 prompt is below.

\textbf{GPT-4 Prompt: }\textit{Please construct a realistic, approximately \{5, 10\}-turn dialogue directly involving a \{2, 5, 10, 15\}-year-old \{toddler, child, teenager\}\footnote{\textit{toddler} is used for age 2, \textit{child} for ages 5 and 10, and \textit{teenager} for age 15.} as a participant. The \{toddler, child, teenager\} is the central participant in the dialogue, with most/all speech directed towards them. Hence, for this dialogue, please limit the vocabulary to that of which a typical \{2, 5, 10, 15\}-year-old \{toddler, child, teenager\} would understand. The dialogue should be \{type\}.\footnote{A random conversation type along with its explanation is sampled each time from the ones in Table \ref{tab:TD_conversation_types}.} The dialogue should use the verb `\{verb\}', the noun `\{noun\}', and the adjective `\{adjective\}'. Please include the following participants along with the child: \{participants\}.\footnote{If $turn = 5$, we randomly sample one additional participant from the corresponding list in Table \ref{tab:TD_participants}. For $turn = 10$, we randomly sample two additional participants.} Participant labels should be surrounded by double asterisks, i.e. `**participant**'. If there are several of the same type of participant (e.g. multiple friends or classmates), please label them distinctly, e.g. `**Friend 1**' and `**Friend 2**'. Please list and describe the participants after `PARTICIPANTS:', briefly describe the context/setting of the dialogue after `SETTING:', and present the dialogue itself after `DIALOGUE:'. The turns of the dialogue should be separated by `\escape{n}\escape{n}'. Remember, please ensure the dialogue is realistic, and one that would likely occur in the real world directly involving a \{2, 5, 10, 15\}-year-old \{toddler, child, teenager\}."
}

\begin{table*}[t]
\centering
\small
\begin{tabularx}{\textwidth}{p{0.5cm}|p{14.5cm}}
\toprule
\textbf{Age} & \textbf{Example}\\
\midrule
2 & **Babysitter**: Look, the pool is all empty. All gone! \escape{n}\escape{n} **Toddler**: Water all gone? \escape{n}\escape{n} **Babysitter**: Yes, all gone. We let the water out. It went down, down, down. \escape{n}\escape{n} **Toddler**: Why? \escape{n}\escape{n} **Babysitter**: So the pool can dry. We don't want it to get yucky. \escape{n}\escape{n} **Toddler**: Pool fall? \escape{n}\escape{n} **Babysitter**: No, the pool didn't fall. The water fell out when we opened the plug. \escape{n}\escape{n} **Toddler**: Plug? \escape{n}\escape{n} **Babysitter**: Yes, the plug is like a little stopper that keeps the water in. When we pull it, water comes out. \escape{n}\escape{n} **Toddler**: More water? \escape{n}\escape{n} **Babysitter**: Not now. We'll fill the pool again another day, okay? \escape{n}\escape{n} **Toddler**: Okay. Play ball? \escape{n}\escape{n} **Babysitter**: Sure, let's play ball now. The pool will be here when we're ready to fill it up again.\\
\midrule
5 & **Teacher**: "Alright, everyone, it's time to clean up! **Child**, can you please help me by putting the crayons back in the box?" \escape{n}\escape{n} **Child**: "Yes! I can do that. The box is empty so I'll fill it up!" \escape{n}\escape{n} **Teacher**: "Thank you, that's very helpful. Make sure the lids are on tight so they don't dry out." \escape{n}\escape{n} **Child**: "I did it! Look, they're all inside now." \escape{n}\escape{n} **Teacher**: "Great job! Now, **Classmate**, can you and **Child** go and put the books back on the shelf, please?" \escape{n}\escape{n} **Classmate**: "Can we go to play after?" \escape{n}\escape{n} **Teacher**: "Yes, as soon as our classroom is tidy, we can all go outside and play."  \escape{n}\escape{n} **Child**: "Come on, let's put the books away fast so we can go play!"  \escape{n}\escape{n} **Classmate**: "Okay, let's do it! Race you to the shelf!"  \escape{n}\escape{n} **Teacher**: "Be careful running, but I love the enthusiasm! Thank you both for helping."\\
\midrule
10 & **Dad**: "Once upon a time, in a faraway kingdom, there lived an earless rabbit who loved to make pancakes." \escape{n}\escape{n} **Child**: "An earless rabbit? How could he hear if he wanted to flip the batter?" \escape{n}\escape{n} **Dad**: "Well, you see, this rabbit had a special talent. He could feel the vibrations of the batter sizzling on the pan. When it was time to flip, he'd give it a perfect toss." \escape{n}\escape{n} **Child**: "That's so cool! Did the rabbit have any friends?" \escape{n}\escape{n} **Dad**: "Yes! His best friend was a turtle who loved to swim. One day, they decided to have a pancake picnic by the lake." \escape{n}\escape{n} **Child**: "Did they swim in the lake after eating pancakes?" \escape{n}\escape{n} **Dad**: "They sure did. The turtle taught the rabbit how to swim, and they had the best day splashing around until the sun set."\\
\midrule
15 & **Girlfriend**: Hey, so what's the plan for this history project video? \escape{n}\escape{n} **Teenager**: We need to make a mini-documentary about the industrial revolution. I was thinking we could start by showing how machines changed production, like how they used to churn butter by hand before. \escape{n}\escape{n} **Girlfriend**: Oh, cool idea! We could use that old butter churn in your grandma's attic for a visual. What role do you want me to play in the video? \escape{n}\escape{n} **Teenager**: Could you narrate the parts about the technological advancements? You're really good at explaining stuff. \escape{n}\escape{n} **Younger Sibling**: Can I help too? I want to be in the video! \escape{n}\escape{n} **Teenager**: Sure, you can help us set up the scenes. But no forcible taking over, okay? We need to work together as a team. \escape{n}\escape{n} **Younger Sibling**: I promise I'll be good! Can I churn the butter for the scene? \escape{n}\escape{n} **Teenager**: That's perfect! It'll look more authentic with you doing it. Let's get everything ready and start filming. Thanks for helping out, both of you.\\
\bottomrule
\end{tabularx}
\caption{Examples of collected TinyDialogues conversations by seed age.}
\label{tab:TD_examples}
\end{table*}

\begin{table*}[t]
\centering
\resizebox{\textwidth}{!}{
\begin{tabular}{c|c}
\toprule
\textbf{Conversation Type} & \textbf{Explanation}\\
\midrule
Explanatory & It should involve explaining something(s) and potentially answering question(s).\\
Functional & It should involve attempting to get something(s) done or accomplishing particular goal(s).\\
Narrative & It should involve telling a story (real or fictional) or sharing/recounting an experience.\\
Argumentative & It should involve conflict(s) or disagreement(s) that lead to an argument.\\
& In most cases, the argument should be resolved, resulting in the \{child, toddler, teenager\} learning.\\
\bottomrule
\end{tabular}
}
\caption{The four TinyDialogues conversation types along with their explanations.}
\label{tab:TD_conversation_types}
\end{table*}

\begin{table*}[t]
\centering
\small
\begin{tabular}{c|c}
\toprule
\textbf{TD Seed Age} & \textbf{Potential Participants}\\
\midrule
2 & \{mom, dad, older sibling, babysitter\}\\
\midrule
5 & \{mom, dad, older sibling, younger sibling, teacher,\\
& friend, classmate, grandparent, babysitter, neighbor\}\\
\midrule
10 & \{mom, dad, older sibling, younger sibling, teacher,\\
& friend, classmate, grandparent, babysitter, neighbor\}\\
\midrule
15 & \{mom, dad, older sibling, younger sibling, teacher, friend, classmate, \\
& grandparent, neighbor, coach, tutor, boyfriend, girlfriend\}\\
\bottomrule
\end{tabular}
\caption{The list of other potential participants in each TinyDialogues conversation by seed age.}
\label{tab:TD_participants}
\end{table*}

\section{Data Format \& Preprocessing}
\label{appendix:data_preprocessing}

\textbf{CHILDES: } We noticed some duplicate utterances in CHILDES conversations and removed these to the best of our ability. We preprocessed the CHILDES data to match the format of the TD examples in Table \ref{tab:TD_examples}. See below for an example of part of a single training example for CHILDES. Double asterisks surround speaker labels, double newline tokens separate utterances, and an end-of-text token marks the end of the conversation. Hence, this format was consistent across all conversations in both CHILDES and TD datasets.

\textit{**CHI**: Are those your stars? \escape{n}\escape{n} **MOT**: Can you say star? \escape{n}\escape{n} **CHI**: Star. \escape{n}\escape{n} **CHI**: Look. \escape{n}\escape{n} **CHI**: Stars. \escape{n}\escape{n} **MOT**: Stars. See? Look, look at the yellow star, a golden star. <|endoftext|>}

\textbf{TinyDialogues:} One inconsistency in the collected TD data was that the speaker label for the target child varied across conversations and ages. For 2-year-olds, GPT-4 labeled the child \textit{toddler}, and 15-year-olds were labeled \textit{teenager}. This is likely due to our prompt as discussed in Appendix \ref{appendix:TD_collection}. Further, within the same age, sometimes the label also differed (e.g. \textit{Child}, \textit{5-year-old child}, \textit{5-year-old}). To align with CHILDES, which only used the speaker label CHI for every target child, and make Zorro evaluation consistent across the board (see Appendix \ref{appendix:zorro}), we converted several plausible target child speaker labels in our TD dataset (based on manual examination) to \textit{Child}. We also tried our best to correct any other issues in the GPT-4 outputs, including occasional inconsistencies with newlines and double newline tokens.

\textbf{BabyLM, Wikipedia, \& OpenSubtitles:} For our BabyLM dataset, we concatenated the data across the BabyLM sub-datasets, then sampled approximately 29M words to match the amount of data in CHILDES and TD, while keeping the original distribution among its sub-datasets intact. We sampled in order (i.e. starting from the beginning of each sub-dataset), as several BabyLM examples (e.g. for Wikipedia) span multiple lines, and random sampling would have broken up individual examples and eliminated coherence. We do no further preprocessing to the BabyLM data and keep the format of the original sub-datasets. In particular, we do not add an <|endoftext|> token to the end of each example (like we do with CHILDES and TD) as it is unclear where each example ends. We preprocessed the data for Wikipedia and OpenSubtitles in a very similar fashion to BabyLM. For Wikipedia, we sample a mix of $\approx$ 12M Wikipedia and 17M Simple Wikipedia tokens.

The Python NLTK package's \textit{word\_tokenize} function was used as part of our statistics calculations (discussed in Appendix \ref{appendix:data_statistics}). The parameters for this function are: \textit{language} is `english' (default) and \textit{preserve\_line} is `False' (default) so line breaks are ignored. Specifically, it was used for calculating the number of unique words in Appendix \ref{appendix:data_statistics}. For consistency purposes, data processing and sampling, and other word-related statistics (e.g. total word count, avg. words per utterance) were done by simply splitting the text by space.

\section{Dataset Statistics}
\label{appendix:data_statistics}

\begin{figure}[t]
    \centering
    \includegraphics[width=0.50\textwidth]{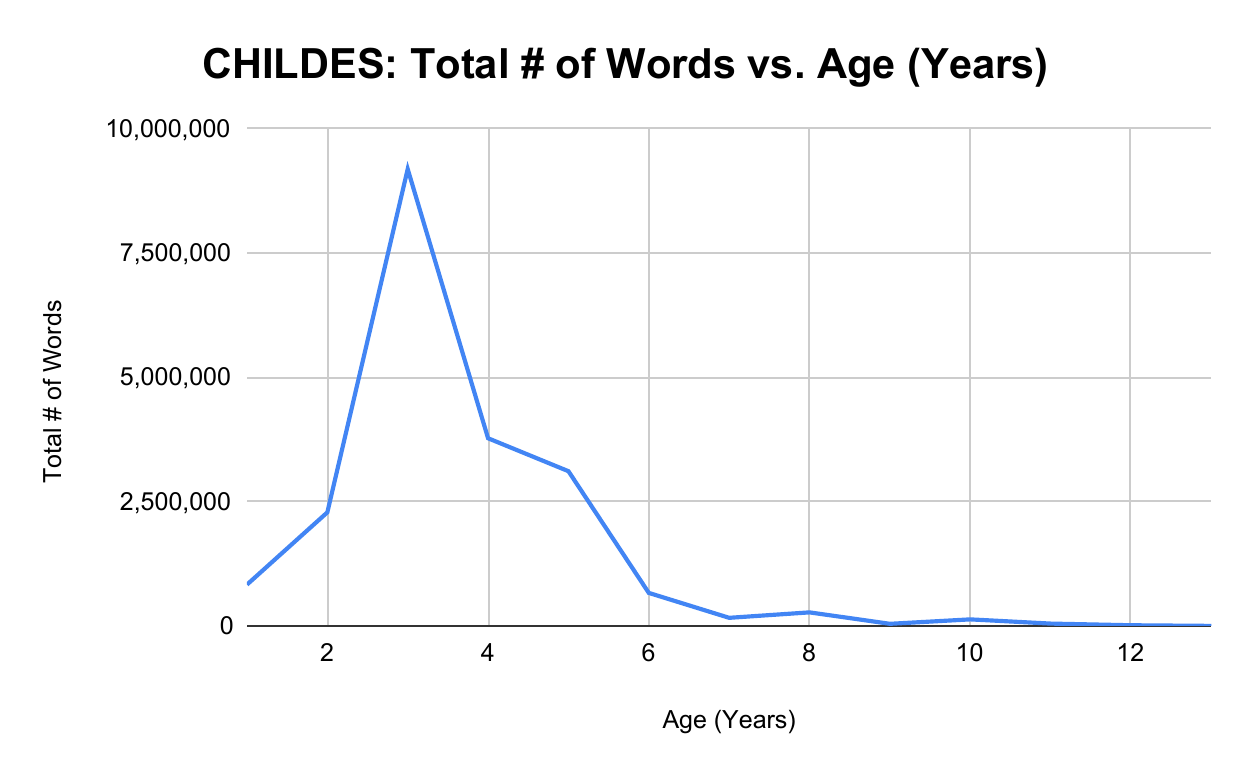}
    \caption{Total CHILDES word counts (utterances only, no metadata) by age.} \label{fig:CHILDES_words_skew_graph}
\end{figure}

\textbf{CHILDES} consists of $\approx$11k conversations over $\approx$29M words. The mean length of utterances is low, at only 3.85 words (minus speaker label), which is likely partially due to the skew in age, where $\approx$90\% of the data is for children ages 2-5 (see Figure \ref{fig:CHILDES_words_skew_graph}). CHILDES contains $\approx$74.5k unique words and $\approx$448 utterances (on avg.) per conversation.

\textbf{BabyLM, Wikipedia, \& OpenSubtitles:} Our BabyLM dataset consists of $\approx$443k unique words in the $\approx$29M word subsample we take for our experiments. Our Wikipedia and OpenSubtitles datasets contain $\approx$644k and 301k unique words, respectively. Individual example statistics are not available as many or all examples span multiple lines, and no end-of-example markers were given. More specific details and statistics about the BabyLM dataset (including its sub-datasets) can be found in \citet{warstadt2023findings}, e.g. Table 1 in their paper.

\textbf{TinyDialogues} consists of $\approx$130k conversations across $\approx$29M words. There are $\approx$14 utterances (on avg.) per conversation, $\approx$96k unique words, and 13.42 words (on avg.) per utterance (minus speaker label). Since TD contains a uniform distribution across ages (including older ages), it is not surprising that the word diversity and average length of utterance are higher than CHILDES. Further, the average TD conversation is shorter than CHILDES, resulting in a higher number of individual conversations. More detailed statistics for TD (broken down by age) are in Table \ref{tab:TD_dataset_stats}. As expected, the vocabulary (unique words) and average length of utterance increase with age. Conversely, the total number of conversations and average utterances per conversation decrease with age.

\begin{table*}[t]
\centering
\small
\begin{tabular}{cccccc}
\toprule
\textbf{Age} & \textbf{Conversations} & \textbf{Total Words} & \textbf{Unique Words} & \textbf{Avg. Utterances per Convo} & \textbf{Avg. Words per Utterance}\\
\midrule
2                 & 43,697 & 7,183,704 & 16,269 & 15.75 & 8.32\\
5                 & 33,248 & 7,194,902 & 15,534 & 14.01 & 13.36\\
10                & 27,198 & 7,196,356 & 42,508 & 13.61 & 17.35\\
15                & 25,589 & 7,199,752 & 73,484 & 12.88 & 19.77\\
Total             & 129,732 & 28,774,714 & 95,868 & 14.29 & 13.42\\
\bottomrule
\end{tabular}
\caption{TinyDialogues dataset statistics broken down by seed age.}
\label{tab:TD_dataset_stats}
\end{table*}

\section{Further Training \& Compute Details}
\label{appendix:compute_details}

We searched through different values of the learning rate (LR) for GPT-2 training. Specifically, $LR = \{1e-06,5e-06,1e-05,5e-05,1e-04,5e-04,1e-03\}$. Through initial experiments, we found that $LR = 1e-04$ seemed to result in the best convergence behavior across the board, and used that for all our training experiments. We do the same for RoBERTa, searching through $LR = \{5e-06,2e-05,5e-05,1e-04,5e-04\}$, and choose $LR = 5e-05$ for all experiments.

Our experiments were run on varying GPUs. This included a single RTX 3090TI (24GB VRAM), up to eight A40s (48GB VRAM each), up to eight A100s (80GB VRAM each), and up to four H100s (80GB VRAM each). Training time varied by the type and number of GPUs and the particular experiment (e.g. number of repeated buckets).

\section{Zorro Evaluation Details}
\label{appendix:zorro}

Zorro was inspired by BLiMP \cite{warstadt-etal-2020-blimp-benchmark} and is a modification for child-directed language (e.g. lower vocabulary). However, it was designed specifically for masked language models such as RoBERTa. To adapt it to GPT-2, we reformatted the Zorro data to match the BLiMP format and used the BLiMP evaluation in the BabyLM evaluation suite\footnote{\url{https://github.com/babylm/evaluation-pipeline-2023}} since the difference is mainly just the data. Further, we use the full Zorro test suite and do not filter examples by vocabulary. Hence, our results are not comparable to \citet{qin2024systematic} who filter Zorro examples by training vocabulary.

To better match the training data format and assess the effects of speaker labels, we came up with three variations of Zorro: 1) the original Zorro sentences (used to assess BabyLM), 2) the sentences with a common CHILDES speaker label prepended (used to assess CHILDES), and 3) the sentences with a common TD speaker label prepended (used to assess TD). To further match the training data as shown in Appendix \ref{appendix:data_preprocessing}, the speaker labels were surrounded by double asterisks, and sentences included double newline tokens (before and after). We used the mother speaker label (MOT) for CHILDES, and the child speaker label (Child) for TD (see Appendix \ref{appendix:data_preprocessing}), as these were some of the most frequent speaker labels for each dataset respectively (see Table \ref{tab:speaker_labels}). Further, preliminary experiments showed that these particular labels worked best for assessing each model.

\begin{table}[t]
\centering
\small
\begin{tabular}{lccc}
\toprule
\textbf{Dataset} & \textbf{Speaker Label} & \textbf{Frequency} & \textbf{Proportion}\\
\midrule
CHILDES & MOT                 & 1,905,187 & 45.7\%\\
CHILDES & CHI                 & 1,593,073 & 38.2\%\\
CHILDES & INV                 & 188,712 & 4.5\%\\
CHILDES & FAT                 & 164,248 & 3.9\%\\
\midrule
TD & Child               & 735,176 & 46.6\%\\
TD & Mom                 & 132,746 & 8.4\%\\
TD & Dad                 & 129,568 & 8.2\%\\
TD & Older Sibling       & 120,468 & 7.6\%\\
\bottomrule
\end{tabular}
\caption{List of the top speaker labels for CHILDES and TD training splits along with their frequencies and proportions. This is after converting all target child labels for TD to \textit{Child}, as described in Appendix \ref{appendix:data_preprocessing}. For CHILDES: MOT stands for \textit{mother}, CHI for \textit{child}, INV for \textit{investigator}, and FAT for \textit{father}.}
\label{tab:speaker_labels}
\end{table}

\section{Further Experimental Motivation}
\label{appendix:buckets_motivation}

If a dataset can be described as a concatenation of equal-sized buckets A, B, and C, the repeated bucket approach can be described as $An \ Bn \ Cn$. Other than being a compromise between standard iterated training and human learning (as discussed in Section \ref{sec:model_training}), the iterative approach (training across the entire dataset several times) can potentially \textit{wash away} global ordering effects (especially when the epoch count is high) as global order differences mainly exist within each individual epoch. When trained across several epochs, its effects may be less noticeable. The repeated buckets approach maintains the global order across training as a whole. The model can learn more from each bucket before moving to the next. The chosen models for the repeated bucket experiments are the final models at the end of training. 

\section{Statistical Significance for Experimental Results}
\label{appendix:stat_sig}

Statistical significance $p$-values for all the experiments reported in Tables \ref{tab:eval_results_datasets} to \ref{tab:eval_results_local_order_roberta} in Section \ref{sec:results_and_analysis} can be found in Tables \ref{tab:eval_results_datasets_$p$-values} to \ref{tab:eval_results_local_order_roberta_$p$-values}. We use paired two-tailed t-tests, and use $\alpha=0.05$ as the threshold to determine significance. For each experiment, we compare across the ordered concatenation of results for all the corresponding seeds of each model.

For Zorro, we calculate statistical significance by comparing the individual per-example results of each model (and seed), e.g. if the model answered the particular example correctly (1) or not (0). Since the metric for WS is correlation, it is not feasible to break this down to a per-example level. As such, we instead compare across the correlation scores of the models on the individual WS sub-datasets, namely RG-65, WordSim-353, SimLex-99, SimVerb-3500, and MTest-3000.

\begin{table}[t]
\centering
\scalebox{0.82}{
\begin{tabular}{cccc}
\toprule
\textbf{Model A} & \textbf{Model B} & \textbf{$p$-value} & \textbf{Significant?}\\
\midrule
CHILDES                   & \textbf{TD} & 2.00E-06 & Yes\\
CHILDES                   & \textbf{Wikipedia} & 0.05 & Yes\\
CHILDES                   & \textbf{OpenSubtitles} & 2.30E-04 & Yes\\
CHILDES                   & \textbf{BabyLM} & 3.11E-05 & Yes\\
\textbf{TD}                   & Wikipedia & 0.01 & Yes\\
TD                   & OpenSubtitles & 0.13 & No\\
TD                   & BabyLM & 0.79 & No\\
\bottomrule
\end{tabular}
}
\caption{Statistical significance $p$-values (using paired two-tailed t-tests) of various pairwise comparisons of our GPT-2 models trained on different datasets. This is for WS, as all Zorro $p$-values were 0 (and hence significant). We use $\alpha=0.05$ to determine significance. We bold the winning model for each significant comparison.}
\label{tab:eval_results_datasets_$p$-values}
\end{table}

\begin{table}[t]
\centering
\scalebox{0.82}{
\begin{tabular}{cccc}
\toprule
\textbf{Model A} & \textbf{Model B} & \textbf{$p$-value} & \textbf{Significant?}\\
\midrule
CHILDES                   & \textbf{TD} & 3.37E-04 & Yes\\
CHILDES                   & \textbf{Wikipedia} & 0.01 & Yes\\
CHILDES                   & OpenSubtitles & 0.17 & No\\
CHILDES                   & \textbf{BabyLM} & 0.01 & Yes\\
TD                   & Wikipedia & 0.29 & No\\
TD                   & OpenSubtitles & 0.17 & No\\
TD                   & BabyLM & 0.57 & No\\
\bottomrule
\end{tabular}
}
\caption{Statistical significance $p$-values (using paired two-tailed t-tests) of pairwise comparisons of our RoBERTa models trained on different datasets. This is for WS, as all Zorro $p$-values were 0 (and significant). We use $\alpha=0.05$ to determine significance. We bold the winning model for each significant comparison.}
\label{tab:eval_results_datasets_roberta_$p$-values}
\end{table}

\begin{table}[t]
\centering
\scalebox{0.82}{
\begin{tabular}{ccccc}
\toprule
\textbf{Dataset} & \textbf{Order A} & \textbf{Order B} & \textbf{$p$-value} & \textbf{Significant?}\\
\midrule
CHILDES                   & Age & Reverse & 0.26 & No \\
CHILDES                   & Age & Random & 0.25 & No \\
CHILDES                   & Reverse & Random & 0.09 & No \\
\midrule
TD                   & Age & Reverse & 0.53 & No \\
TD                   & Age & \textbf{Random} & 0.02 & Yes\\
TD                   & Reverse & \textbf{Random} & 0.02 & Yes\\
\bottomrule
\end{tabular}
}
\caption{Statistical significance $p$-values (using paired two-tailed t-tests) of pairwise comparisons of our GPT-2 models for different global ordering methods, broken down by dataset. This is for WS, as all Zorro $p$-values were 0 (and significant). We use $\alpha=0.05$ to determine significance. We bold the winning global order for each significant comparison.}
\label{tab:eval_results_global_order_$p$-values}
\end{table}

\begin{table}[t]
\centering
\scalebox{0.58}{
\begin{tabular}{ccccccc}
\toprule
\textbf{Dataset} & \textbf{Order A} & \textbf{Order B} & \textbf{Zorro $p$-value} & \textbf{Sig?} & \textbf{WS $p$-value} & \textbf{Sig?}\\
\midrule
CHILDES              & \textbf{Normal} & Random & 0 & Yes & 5.97E-04 & Yes \\
TD                   & Normal & Random & 0.31 & No & 0.42 & Yes \\
\bottomrule
\end{tabular}
}
\caption{Statistical significance $p$-values (using paired two-tailed t-tests) of normal vs. random local ordering of our GPT-2 models. We use $\alpha=0.05$ to determine significance. We bold the winning local order for each significant comparison.}
\label{tab:eval_results_local_order_$p$-values}
\end{table}

\begin{table}[t]
\centering
\scalebox{0.58}{
\begin{tabular}{ccccccc}
\toprule
\textbf{Dataset} & \textbf{Order A} & \textbf{Order B} & \textbf{Zorro $p$-value} & \textbf{Sig?} & \textbf{WS $p$-value} & \textbf{Sig?}\\
\midrule
CHILDES              & \textbf{Normal} & Random & 0 & Yes & 2.42E-04 & Yes \\
TD                   & \textbf{Normal} & \textbf{Random} & 0 & Yes & 3.75E-03 & Yes \\
\bottomrule
\end{tabular}
}
\caption{Statistical significance $p$-values (using paired two-tailed t-tests) of normal vs. random local ordering of our RoBERTa models. We use $\alpha=0.05$ to determine significance. We bold the winning local order for each significant comparison. Note that for TD, Normal order wins for WS but Random order wins for Zorro.}
\label{tab:eval_results_local_order_roberta_$p$-values}
\end{table}

\section{Further Experimental Results}
\label{appendix:results}

In addition to the experiments discussed in Section \ref{sec:results_and_analysis}, we also report RoBERTa global ordering experiment results in Table \ref{tab:eval_results_global_order_roberta}. As discussed in Section \ref{sec:results_and_analysis}, there were training difficulties, as it appears that the RoBERTa models do not converge properly using our repeated buckets training approach. Hence, they barely achieve above chance on our benchmarks (50\% for Zorro and 0 for WS). 

Furthermore, we tried different values of $n$ (number of times to repeat each bucket) for CHILDES and TD repeated buckets experiments. In particular, $n = 3, 5, 10, 20$. For CHILDES, we also tried different values of $b$ (number of buckets, or approximately equal sections to divide the dataset into) using the global age order. In particular, $b = 3, 5, 10$. We report average results for different values of $n$ and $b$ in Tables \ref{tab:eval_results_n} and \ref{tab:eval_results_b}, respectively. We also compare the typical iterative training approach (20 epochs) to repeated buckets using $n=20$ (analogous to 20 epochs). Results are in Table \ref{tab:eval_results_20-epochs}.

\begin{table}[t]
\centering
\scalebox{0.82}{
\begin{tabular}{lccc}
\toprule
\textbf{Dataset} & \textbf{Order} & \textbf{Zorro} & \textbf{WS}\\
\midrule
CHILDES                 & Age &  54.37\% $\pm$ 2.24\% & 0.02 $\pm$ 0.02\\
CHILDES                 & Reverse &  55.01\% $\pm$ 1.60\% & 0.03 $\pm$ 0.01\\
CHILDES                 & Random & 55.63\% $\pm$ 1.14\% & 0.03 $\pm$ 0.01\\
\midrule
TD                 & Age & 64.43\% $\pm$ 9.18\% & 0.05 $\pm$ 0.05\\
TD                 & Reverse & 56.70\% $\pm$ 1.49\% & 0.02 $\pm$ 0.01\\
TD                 & Random & 57.91\% $\pm$ 2.38\% & 0.02 $\pm$ 0.01\\
\bottomrule
\end{tabular}
}
\caption{Evaluation results (avg. and std. across three seeds) of our RoBERTa models, comparing global ordering methods using the repeated buckets training approach, broken down by dataset. For CHILDES, we use $b=5,n=10$, and for TD, we use $n=5$. As discussed in Section \ref{sec:results_and_analysis}, these models had difficulty converging, and the results are relatively close to random chance.}
\label{tab:eval_results_global_order_roberta}
\end{table}

\begin{table}[t]
\centering
\small
\begin{tabular}{lccc}
\toprule
\textbf{Dataset} & \textbf{$n$} & \textbf{Zorro} & \textbf{WS}\\
\midrule
CHILDES                 & 3 & 68.89\% & 0.10\\
CHILDES                 & 5 & 72.02\% & 0.14\\
CHILDES                 & 10 & 77.01\% & 0.19\\
CHILDES                 & 20 & 75.75\% & 0.23\\
\midrule
TD                 & 3 & 71.51\% & 0.18\\
TD                 & 5 & 74.48\% & 0.23\\
TD                 & 10 & 79.21\% & 0.32\\
TD                 & 20 & 79.65\% & 0.41\\
\bottomrule
\end{tabular}
\caption{Evaluation results (a single seed) of our GPT-2 models, comparing different values of $n$, broken down by dataset. These results are averaged across three different global ordering methods: age order, reverse order, and random order. For CHILDES, we use $b=5$.}
\label{tab:eval_results_n}
\end{table}

\begin{table}[t]
\centering
\small
\begin{tabular}{lccc}
\toprule
\textbf{Dataset} & \textbf{$b$} & \textbf{Zorro} & \textbf{WS}\\
\midrule
CHILDES                 & 3 & 73.36\% & 0.35\\
CHILDES                 & 5 & 72.12\% & 0.35\\
CHILDES                 & 10 & 70.06\% & 0.35\\
\bottomrule
\end{tabular}
\caption{Evaluation results (a single seed) of our CHILDES GPT-2 models, comparing different values of $b$. These results are averaged across three experiments each: global age order with $n=3,5,10$.}
\label{tab:eval_results_b}
\end{table}

\section{Importance of Speaker Labels}
\label{appendix:speaker_labels}

As an additional experiment, we also assess the importance of having speaker labels for each utterance. We train some versions of our models after removing all speaker labels (including their surrounding double asterisks). The results are reported in Table \ref{tab:eval_results_speaker-labels}. As seen, removing speaker labels detriments syntax and grammar learning (Zorro), but semantics (WS) appears unaffected.

\section{Convergence Behavior of GPT-2 Models}
\label{appendix:convergence_behavior}

\begin{figure}[t]
    \centering
    \hspace*{-0.75cm}
    \includegraphics[width=0.50\textwidth]{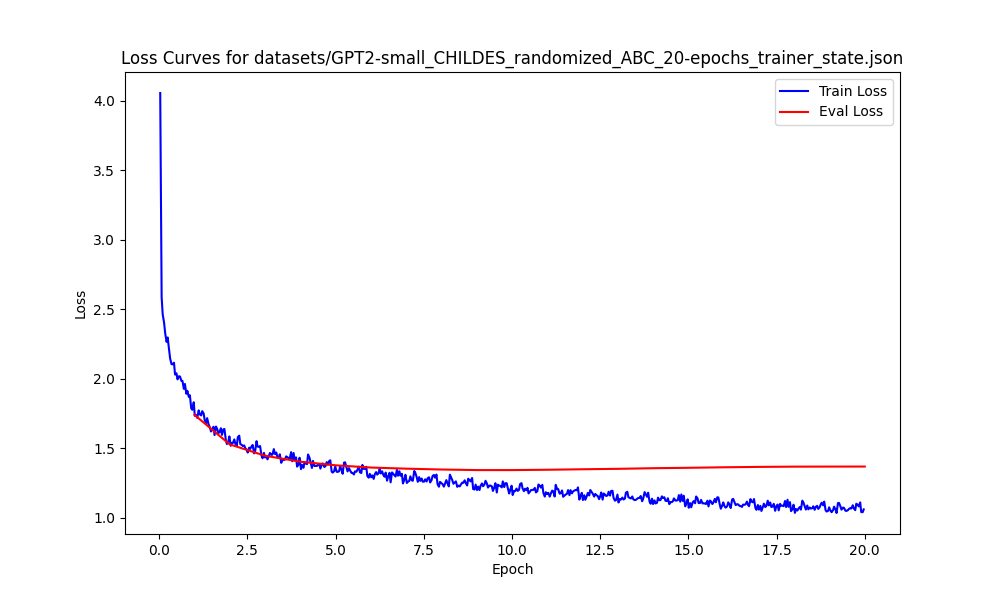}
    \hspace*{-0.75cm} 
    \includegraphics[width=0.50\textwidth]{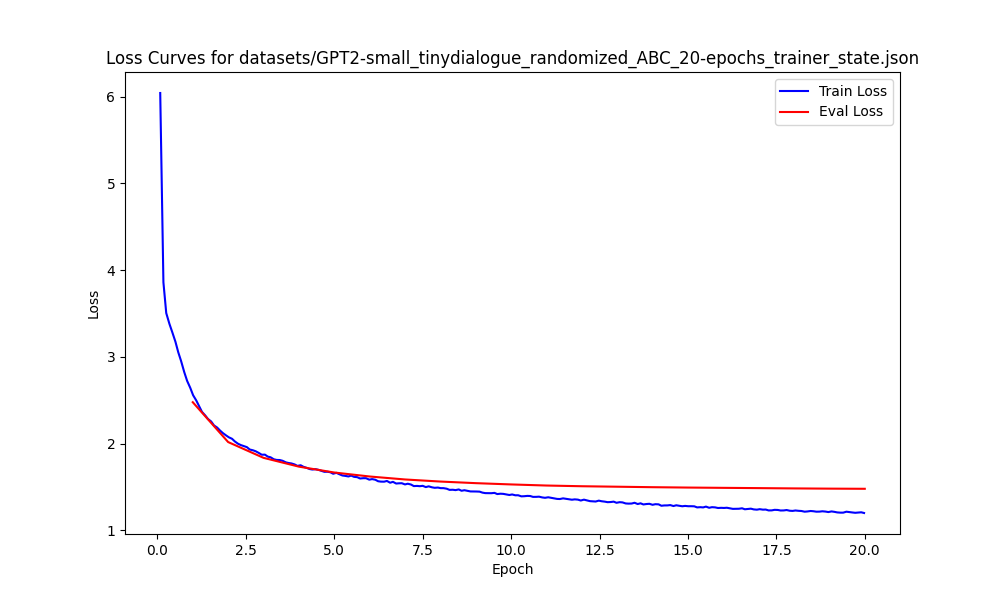}
    \hspace*{-0.75cm} 
    \includegraphics[width=0.50\textwidth]{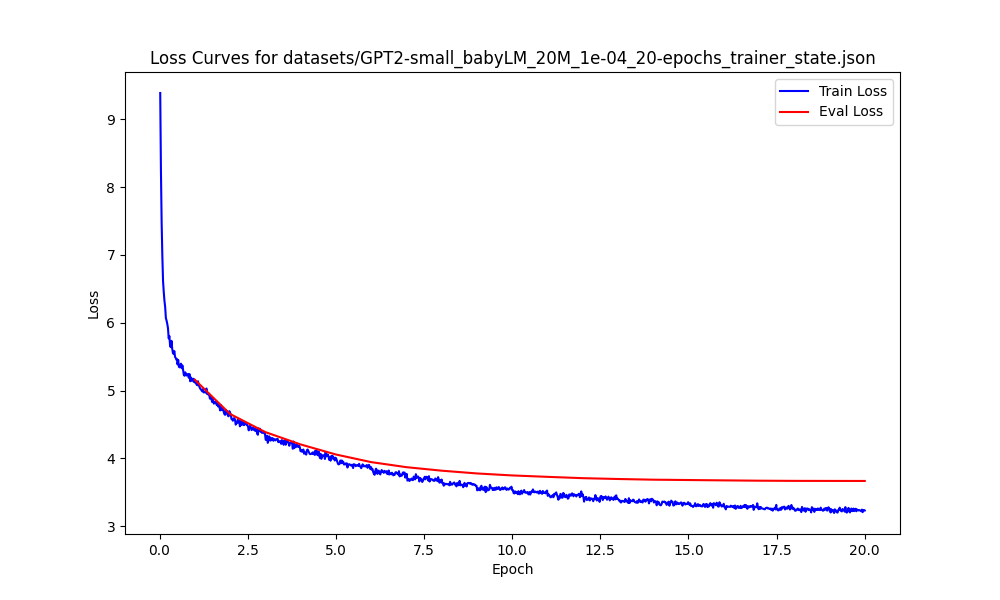}
    \caption{GPT-2 convergence graphs (train and val loss) by dataset, using iterative training for 20 epochs. From top to bottom: CHILDES, TinyDialogues, BabyLM.} \label{fig:dataset_convergence_graphs}
\end{figure}

\begin{figure}[t]
    \centering
    \hspace*{-0.75cm} 
    \includegraphics[width=0.52\textwidth]{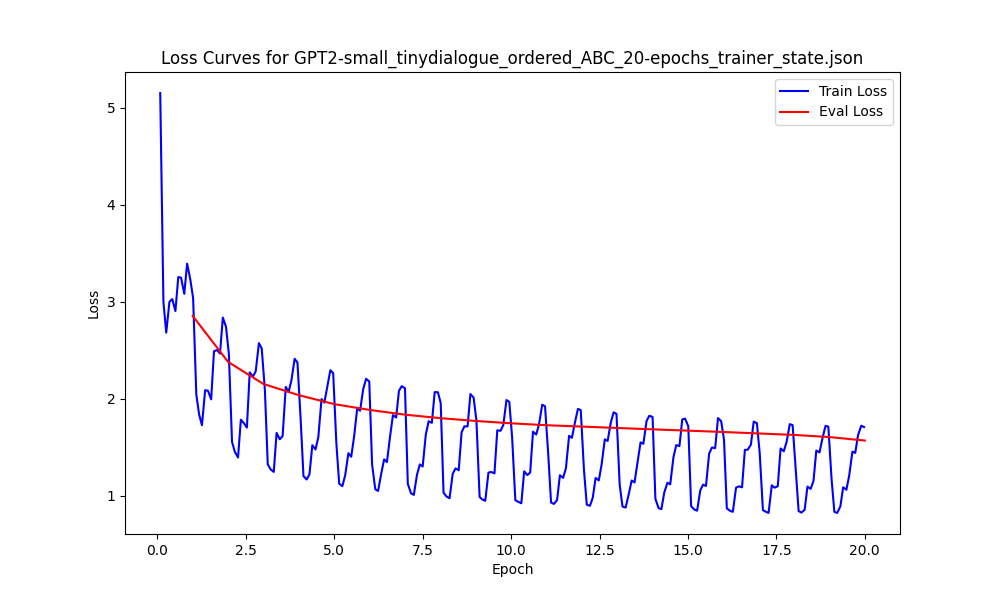}
    \hspace*{-0.75cm} 
    \includegraphics[width=0.52\textwidth]{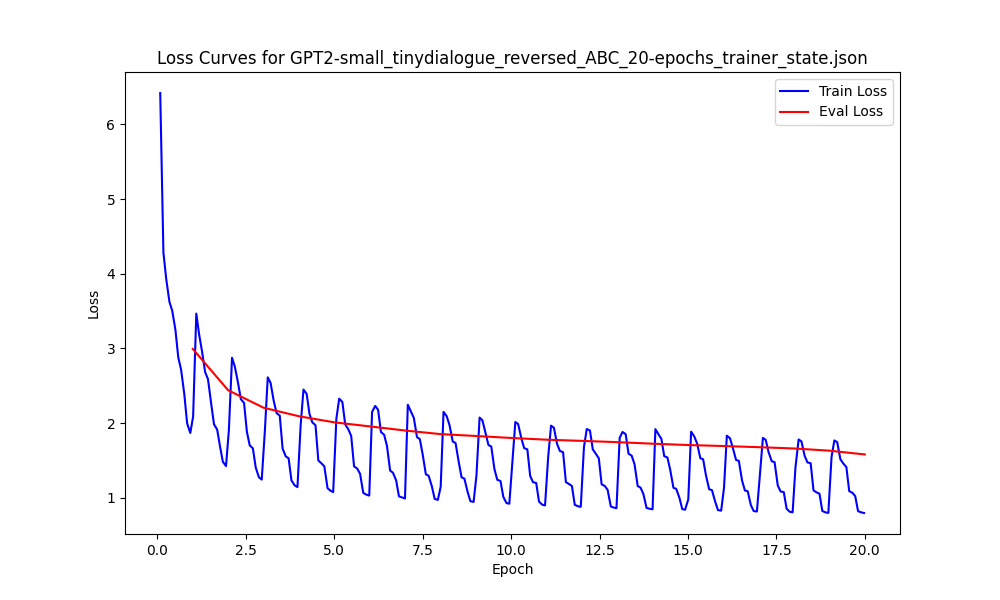}
    \hspace*{-0.75cm} 
    \includegraphics[width=0.52\textwidth]{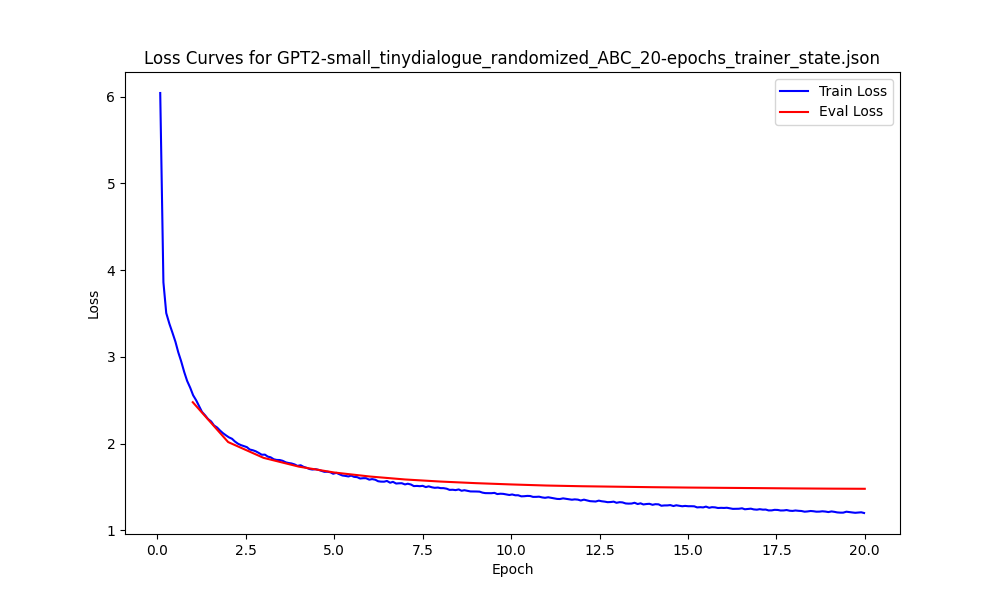}
    \caption{GPT-2 convergence graphs (train and val loss) of TinyDialogues using the typical iterative training approach for 20 epochs, for different global orders. From top to bottom: age order, reverse order, random order.} \label{fig:TD_20-epoch_convergence_graphs}
\end{figure}

\begin{figure}[t]
    \centering
    \hspace*{-0.75cm} 
    \includegraphics[width=0.52\textwidth]{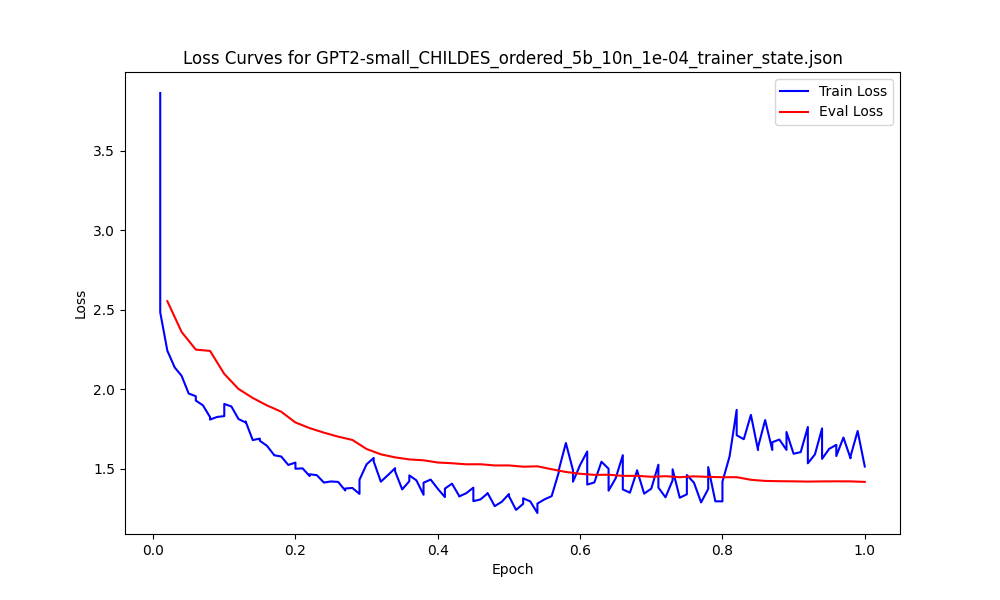}
    \hspace*{-0.75cm} 
    \includegraphics[width=0.52\textwidth]{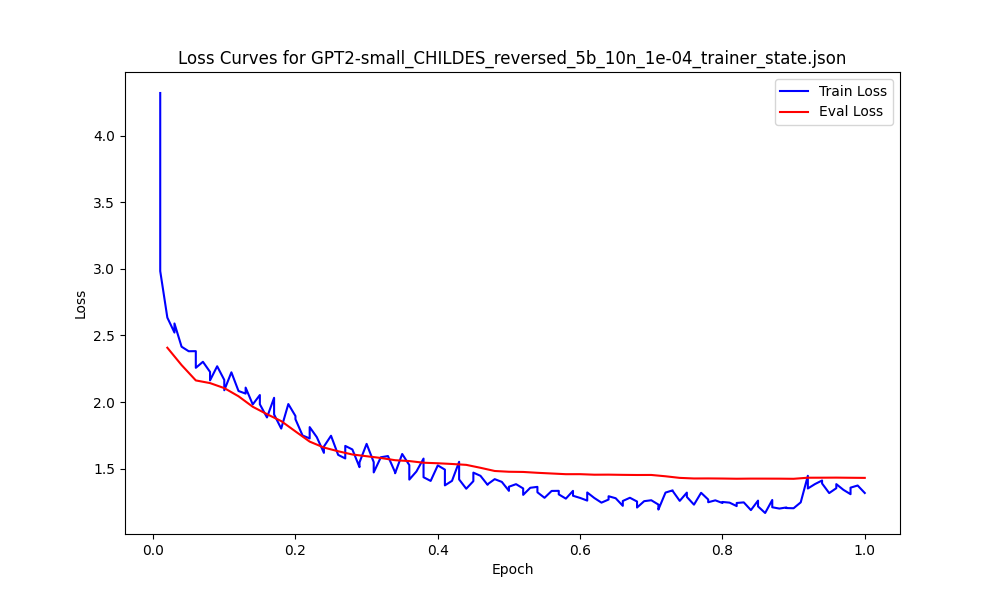}
    \hspace*{-0.75cm} 
    \includegraphics[width=0.52\textwidth]{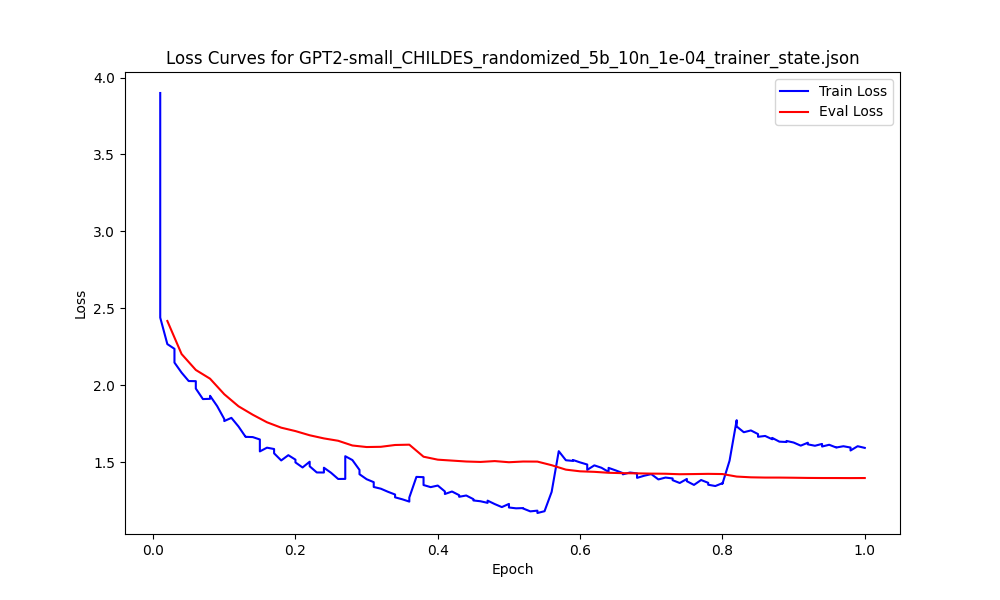}
    \caption{GPT-2 convergence graphs (train and val loss) of CHILDES using the repeated buckets training approach with $b=5, n=10$, for different global orders. From top to bottom: age, reverse, random order.} \label{fig:CHILDES_global_order_10n_convergence_graphs}
\end{figure}

\begin{figure}[t]
    \centering
    \hspace*{-0.75cm} 
    \includegraphics[width=0.52\textwidth]{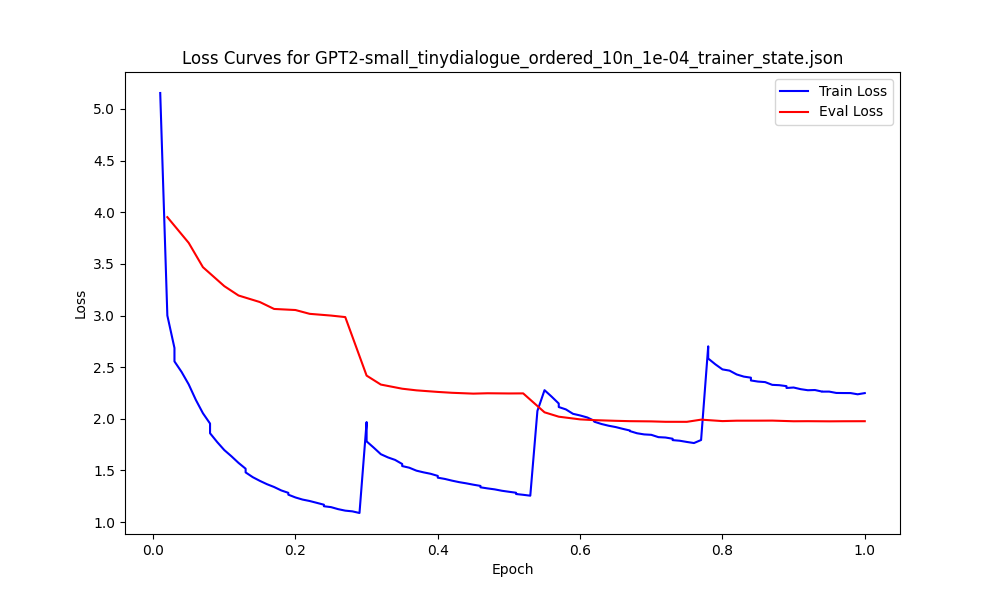}
    \hspace*{-0.75cm} 
    \includegraphics[width=0.52\textwidth]{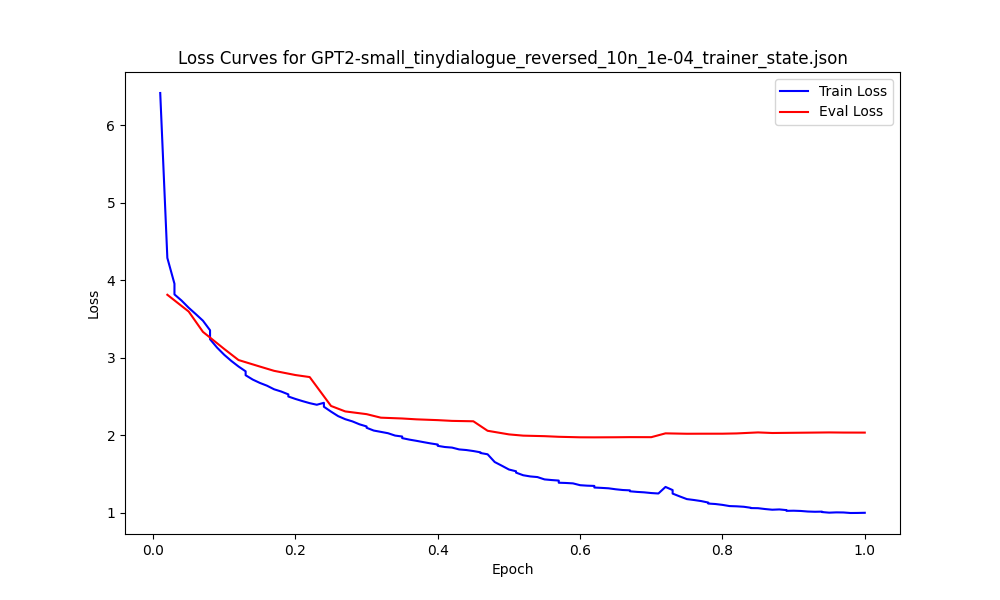}
    \hspace*{-0.75cm} 
    \includegraphics[width=0.52\textwidth]{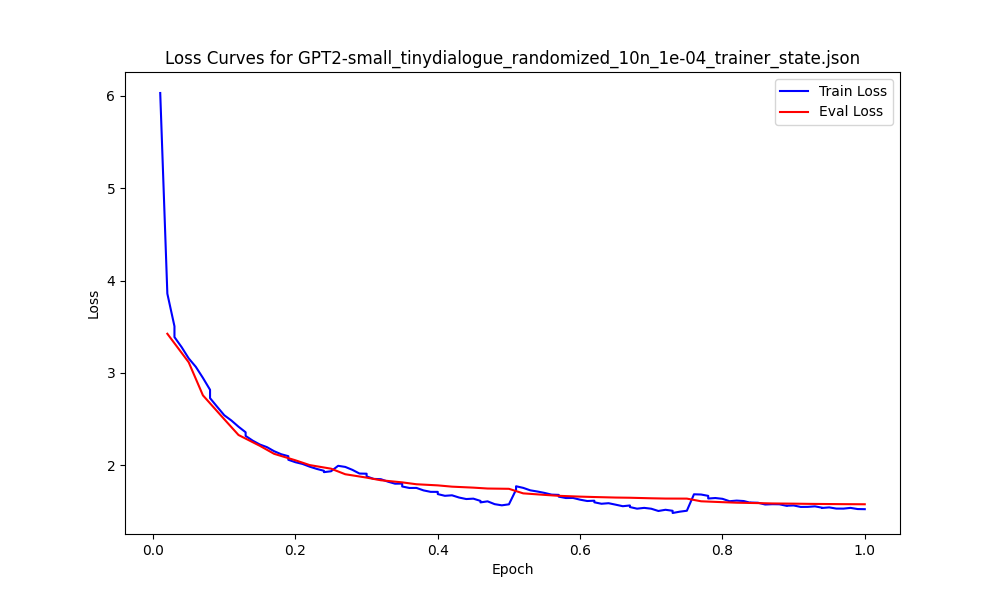}
    \caption{GPT-2 convergence graphs (train and val loss) of TinyDialogues using the repeated buckets training approach with $n=10$, for different global orders. From top to bottom: age order, reverse order, random order.} \label{fig:TD_global_order_10n_convergence_graphs}
\end{figure}

\begin{figure}[t]
    \centering
    \hspace*{-0.75cm} 
    \includegraphics[width=0.50\textwidth]{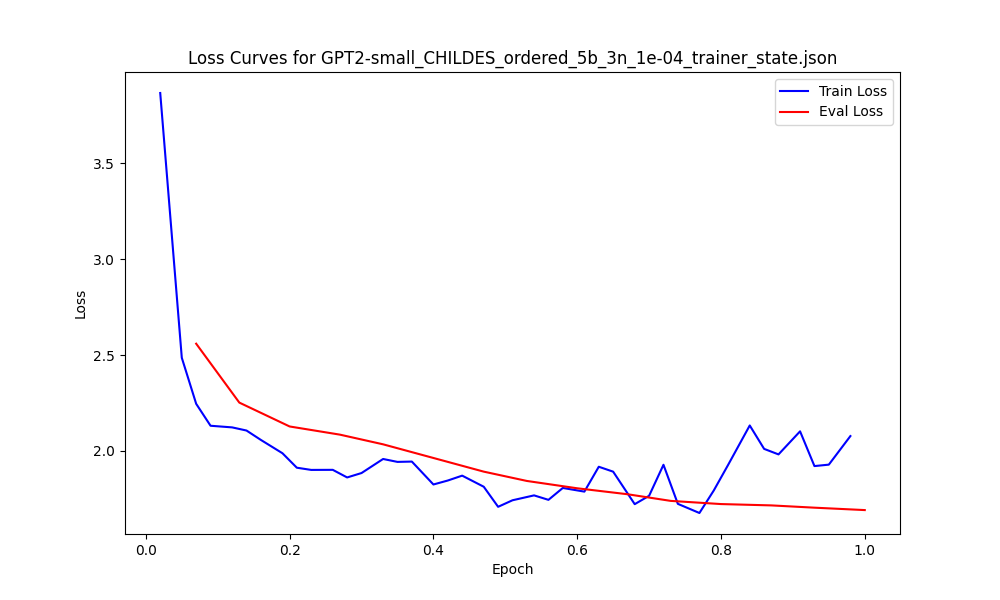}
    \hspace*{-0.75cm} 
    \includegraphics[width=0.50\textwidth]{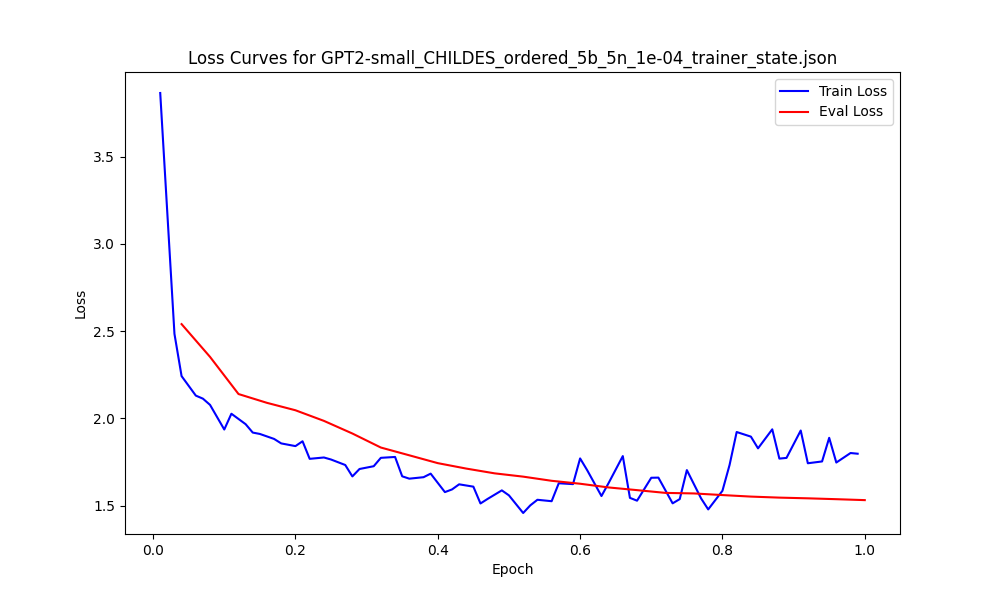}
    \hspace*{-0.75cm} 
    \includegraphics[width=0.50\textwidth]{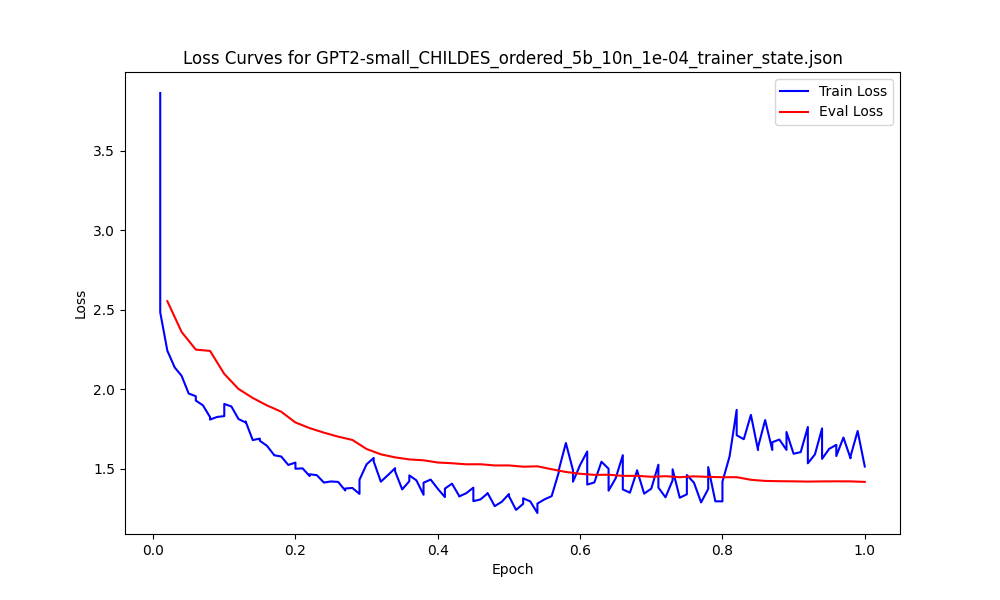}
    \hspace*{-0.65cm} 
    \includegraphics[width=0.50\textwidth]{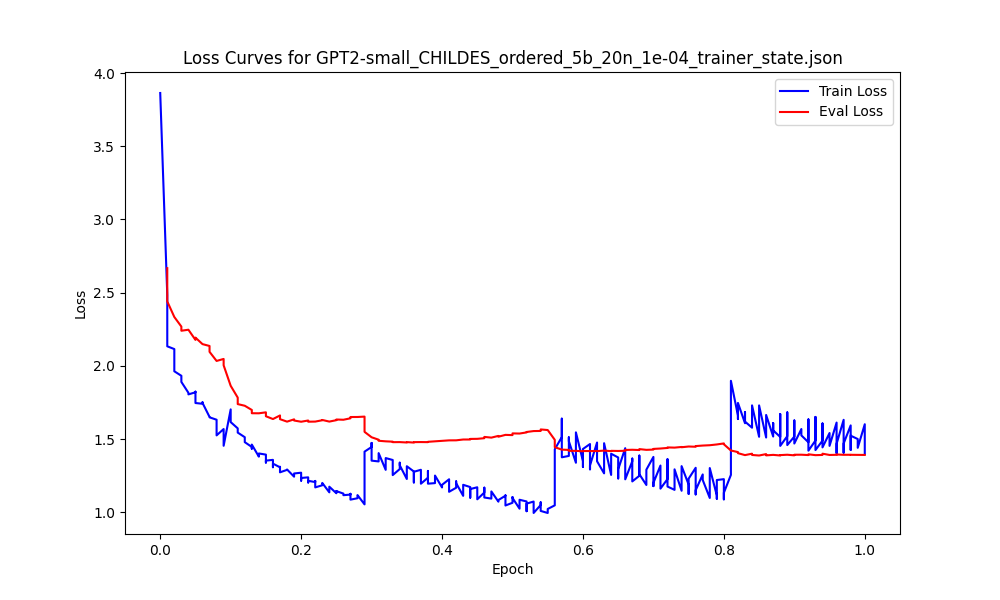}
    \caption{GPT-2 convergence graphs (train and val loss) of CHILDES using the repeated buckets training approach with $b=5$, for different values of $n$. From top to bottom: $n=3,5,10,20$.} \label{fig:bucket_convergence_graphs_by_n}
\end{figure}

\begin{figure}[t]
    \centering
    \hspace*{-0.75cm} 
    \includegraphics[width=0.50\textwidth]{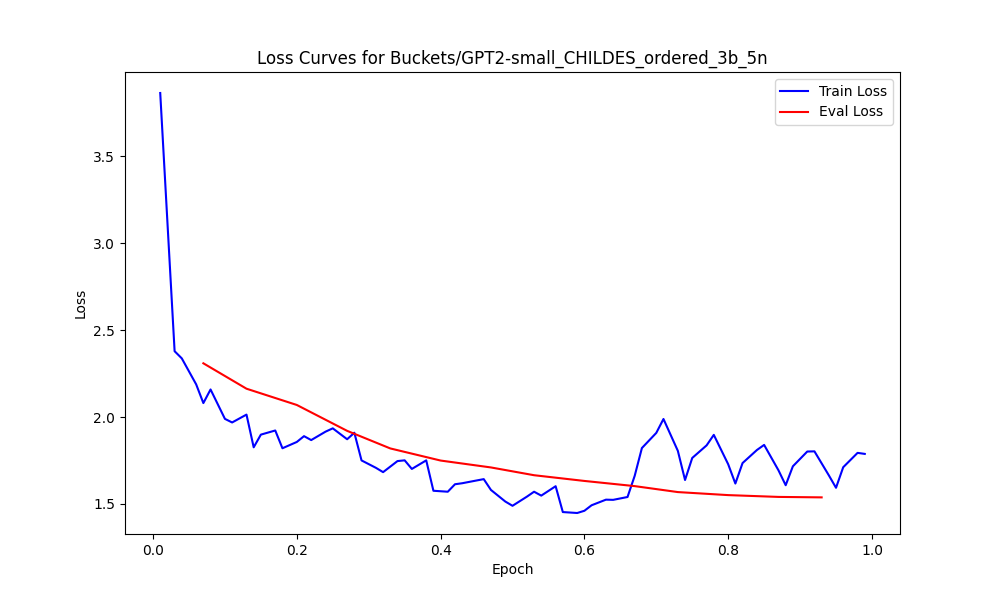}
    \hspace*{-0.75cm} 
    \includegraphics[width=0.50\textwidth]{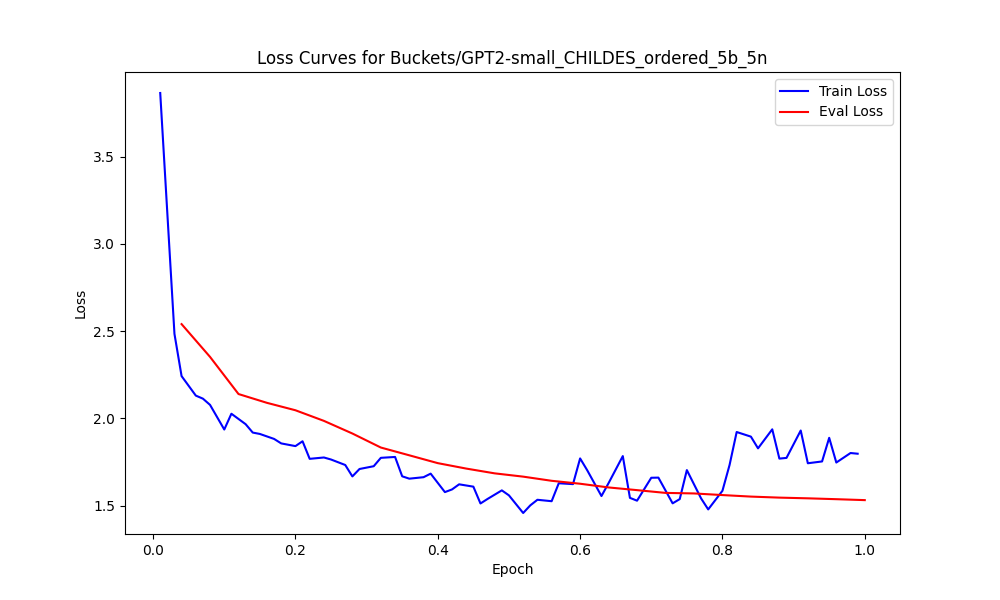}
    \hspace*{-0.75cm} 
    \includegraphics[width=0.50\textwidth]{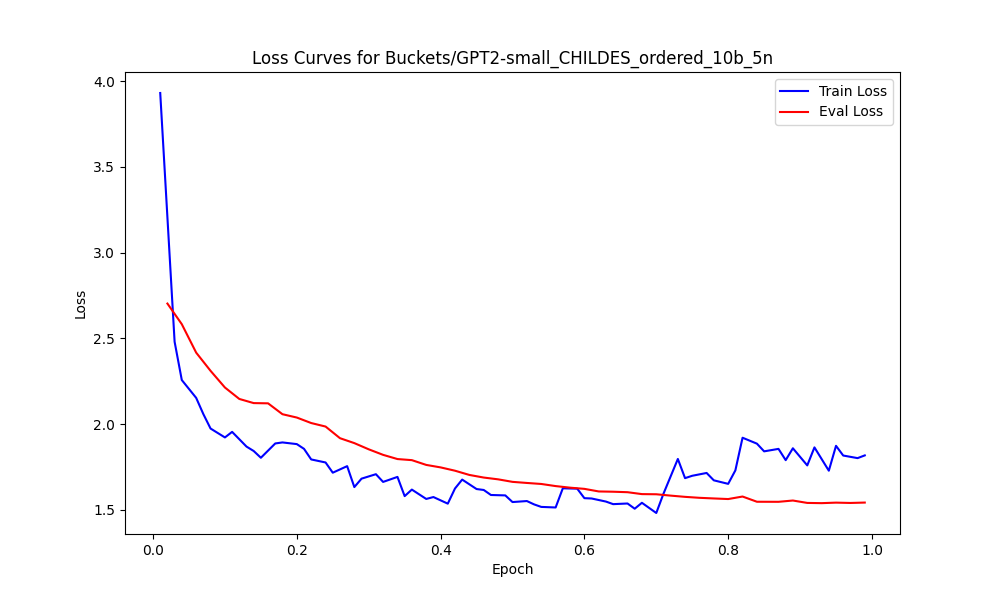}
    \caption{GPT-2 convergence graphs (train and val loss) of CHILDES using the repeated buckets training approach with $n=5$, for different values of $b$. From top to bottom: $b=3,5,10$.} \label{fig:bucket_convergence_graphs_by_b}
\end{figure}

\begin{figure}[t]
    \centering
    \hspace*{-0.75cm}
    \includegraphics[width=0.50\textwidth]{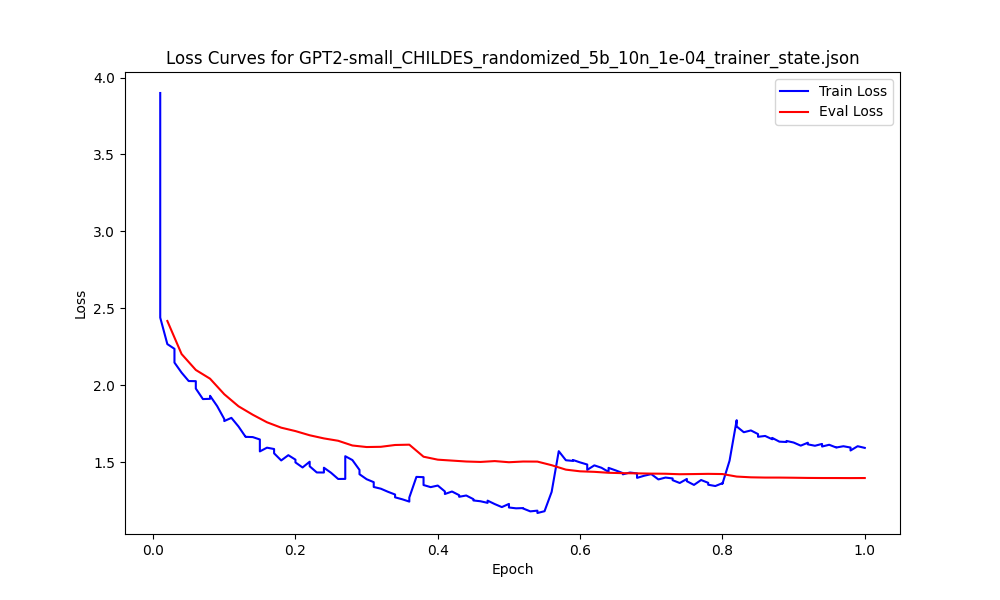}
    \hspace*{-0.75cm}
    \includegraphics[width=0.50\textwidth]{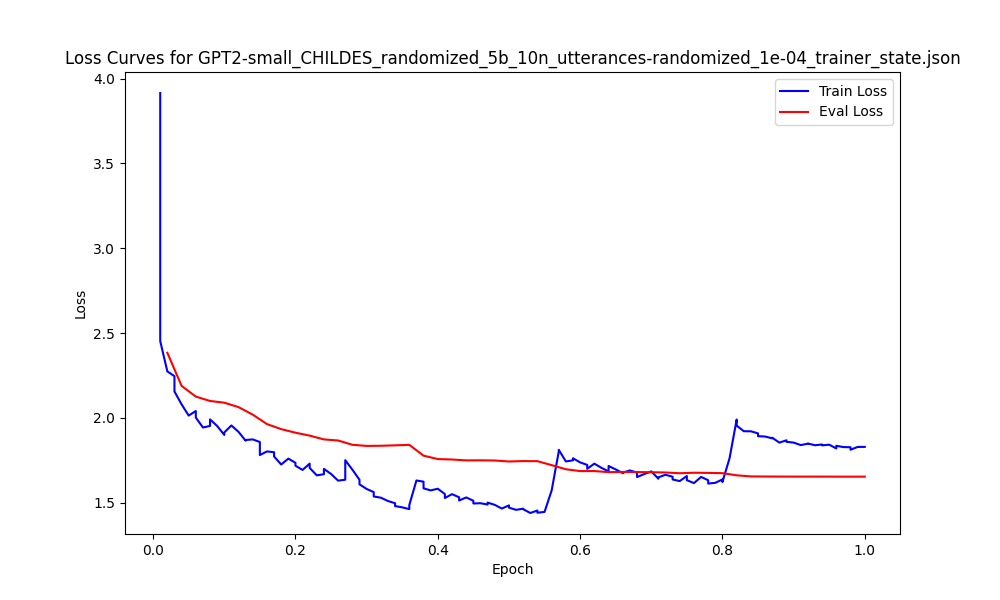}
    \hspace*{-0.75cm}
    \includegraphics[width=0.50\textwidth]{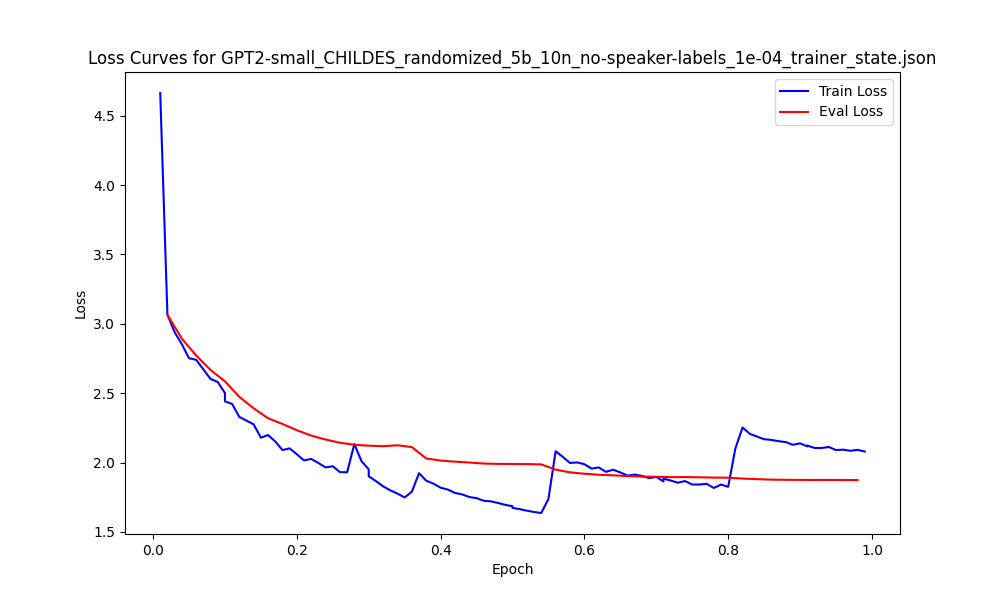}
    \caption{GPT-2 convergence graphs (train and val loss) of CHILDES, looking at the effects of local interventions -- shuffling utterances and removing speaker labels -- using global random ordering with the repeated buckets approach ($b=5,n=10$). From top to bottom: original data (no changes), random shuffling of utterances, no speaker labels for utterances.} \label{fig:micro_intervention_convergence_graphs}
\end{figure}

We plot the convergence graphs (train and validation losses vs. epoch) for several sets of our GPT-2 experiments in Figures \ref{fig:dataset_convergence_graphs} to \ref{fig:micro_intervention_convergence_graphs}. For the repeated buckets experiments, we treat the entire training run as a single epoch. We can notice interesting patterns/trends in the convergence behavior of models depending on several factors including the global ordering and curricularization method. We focus on our GPT-2 experiments as some RoBERTa models did not converge properly (Section \ref{sec:results_and_analysis}).

From Figure \ref{fig:dataset_convergence_graphs}, we see that BabyLM converges to higher losses than CHILDES and TD, although it seems to perform better at test-time for syntax and semantics (as discussed in Section \ref{sec:results_and_analysis}). Losses during training could simply be higher as the dataset is more complicated and varied since it is a mixture.

From Figure \ref{fig:TD_20-epoch_convergence_graphs}, we can see that when we train using the typical iterative epochs approach, the training loss has a cyclical pattern using global age order and reverse order, while it converges smoothly for random order. From Figures \ref{fig:CHILDES_global_order_10n_convergence_graphs} and \ref{fig:TD_global_order_10n_convergence_graphs}, we see that when using the repeated buckets approach for both CHILDES and TD, global age order leads to a slowly cyclical increase in the training loss, while it generally decreases for reverse and random order. Throughout these experiments, while the training loss differs and individual buckets exhibit differing patterns, the high-level behavior and final values of the validation loss, and hence overall learning, are similar. This aligns with the results in Section \ref{sec:results_and_analysis}.

From Figures \ref{fig:bucket_convergence_graphs_by_n} and \ref{fig:bucket_convergence_graphs_by_b}, we see that varying $b$ and $n$ result in minor changes in behavior for the training loss. Specifically, by increasing $n$, the training loss has a more clearly defined cyclical pattern, and the losses converge to lower values. This is expected, since increasing $n$ is analogous to training on more epochs. From Figure \ref{fig:micro_intervention_convergence_graphs}, we see that local interventions -- randomly shuffling utterances and removing speaker labels (see Appendix \ref{appendix:speaker_labels}) -- have minor effects on convergence behavior. However, local interventions result in slightly higher losses overall, especially when removing speaker labels.


\section{Licenses and Intended Use}
\label{appendix:licenses}

We used all existing datasets and models for their intended use. GPT-2 and RoBERTa are licensed under the MIT License. CHILDES is made available under TalkBank which is governed by the Creative Commons CC BY-NC-SA 3.0 copyright license (see \url{https://talkbank.org/share/rules.html}). We plan to release the TinyDialogues dataset under the standard MIT license. Information about the BabyLM challenge and its dataset (which is a collection of portions of several sub-datasets) is at \url{https://babylm.github.io/index.html}.

\begin{table}[H]
\centering
\small
\begin{tabular}{lccc}
\toprule
\textbf{Dataset} & \textbf{Approach} & \textbf{Zorro} & \textbf{WS}\\
\midrule
CHILDES                 & 20-epochs & 77.13\% & 0.52\\
CHILDES                 & $b=5,n=20$ & 75.75\% & 0.48\\
\midrule
TD                 & 20-epochs & 79.41\% & 0.54\\
TD                 & $n=20$ & 79.65\% & 0.54\\
\bottomrule
\end{tabular}
\caption{Evaluation results (a single seed) of our GPT-2 models, comparing typical iterative training (20 epochs) vs. repeated buckets with $n=20$ ($b=5$ for CHILDES), broken down by dataset. These results are averaged across three experiments: the three different global ordering methods (age order, reverse order, random order).}
\label{tab:eval_results_20-epochs}
\end{table}

\begin{table}[H]
\centering
\small
\scalebox{0.82}{
\begin{tabular}{lccc}
\toprule
\textbf{Dataset} & \textbf{Speaker Label?} & \textbf{Zorro} & \textbf{WS}\\
\midrule
CHILDES                 & Yes & 78.29\% $\pm$ 0.51\% & 0.24 $\pm$ 0.01\\
CHILDES                 & No & 76.61\% $\pm$ 1.22\% & 0.24 $\pm$ 0.00\\
\midrule
TD                 & Yes & 78.48\% $\pm$ 0.82\% & 0.42 $\pm$ 0.01\\
TD                 & No & 77.37\% $\pm$ 1.32\% & 0.42 $\pm$ 0.00\\
\bottomrule
\end{tabular}
}
\caption{Evaluation results (avg. and std. across three seeds) of our GPT-2 models, comparing speaker label vs. no speaker label for conversation utterances. We use standard iterative training for 20 epochs. Zorro differences are significant, whereas WS are not, using paired two-tailed t-tests with $\alpha=0.05$ threshold.}
\label{tab:eval_results_speaker-labels}
\end{table}

\section{Code \& Data}
\label{appendix:code_and_data}

All code and data for this project is released at \url{https://github.com/styfeng/TinyDialogues}. Some of the code was written with the assistance of ChatGPT (specifically, GPT-4 and GPT-4o).

\end{document}